%% file: gmas_iclr2022_conference.tex
\documentclass{article} 
\usepackage{gmas_iclr2022_conference,times}

\input{math_commands.tex}

\usepackage{hyperref}
\usepackage{url}
\usepackage{microtype}
\usepackage{graphicx}
\usepackage{amsmath}
\usepackage{booktabs} 
\usepackage{algorithmic}
\usepackage{algorithm}
\usepackage{makecell}
\usepackage{multirow}
\usepackage{threeparttable}
\usepackage{subfigure}
\usepackage{wrapfig,lipsum,booktabs}
\usepackage{graphicx} 
\usepackage{epstopdf}

\title{A Game-Theoretic Approach for Improving Generalization Ability of TSP Solvers}

\author{Chenguang Wang$^{1,*}$, Yaodong Yang$^{2,*}$, Oliver Slumbers$^4$, \\ 
\textbf{Congying Han$^{\dag,1}$, Tiande Guo$^{1}$, Haifeng Zhang$^{3}$, Jun Wang$^{4}$} \\
 $^{1}$University of Chinese Academy of Sciences $^{2}$King's College London \\ $^{3}$Institute of Automation, Chinese Academy of Sciences $^{4}$University College London
\\
\texttt{wangchenguang19@mails.ucas.ac.cn}\\ \texttt{yaodong.yang@kcl.ac.uk}\\ \texttt{hancy@ucas.ac.cn}\\
}

\iclrfinalcopy 
\begin{document}

\maketitle

\begin{abstract}
In this paper, we introduce a two-player zero-sum framework between a trainable \emph{Solver} and a \emph{Data Generator} to improve the generalization ability of deep learning-based solvers for Traveling Salesman Problem (TSP).
Grounded in  \textsl{Policy Space Response Oracle} (PSRO) methods, our two-player framework outputs a  population of best-responding Solvers, over which we can mix and output a combined model that achieves the least exploitability against the Generator, and thereby the most  generalizable performance on different TSP tasks. 
We conduct experiments on a variety of TSP instances with different types and sizes. Results suggest  that our Solvers achieve the state-of-the-art performance even on tasks the Solver never meets, whilst the performance of other deep learning-based Solvers drops sharply due to over-fitting. 
To demonstrate the principle of our framework, we study  the learning outcome of the proposed two-player game and demonstrate that the exploitability of the Solver population decreases during training, and it eventually approximates  the Nash equilibrium along with the Generator.
 \let\thefootnote\relax\footnotetext{$^*$Equal contributions. $^\dag$Corresponding author.} 
\end{abstract}

\section{Introduction}
Deep learning for solving combinatorial optimization problems has recently attracted enormous attention due to its ability to
capture  complex improvement heuristics  from training over millions of problem instances~\citep{dai2017learning}. 
Additionally, due to the efficiency of the forward computation of neural networks, deep learning based techniques are particularly efficient in comparison to traditional methods when performing inference  on large-scale problems. As a consequence, it is promising direction to study training deep learning-based solvers offline  and later to implement solvers online. 



The generalization ability of a solver concerns its performance on a variety of different data distributions.
Most previous works only train their models on data from uniform distribution, however, overfitting to the uniform distribution can cause poor generalization ability. 
As a direct intuition, improvement on the generalization ability can be obtained by exploring the data distribution where the model performs poorly.
In this work, we tackle the generalization problem by introducing a novel two-player game framework: player one, the trainable Solver, aims to train solvers to perform well on distributions chosen by player two, and player two, the Data Generator, aims to generate distributions in which Solver can be challenged. 
With regard to the policy space of Data Generator, it is possible that Data Generator has an infinite-sized number of choices since there are an enormous amount of candidate distributions. In this view, previous work focuses solely on adopting the uniform distribution, which unfortunately is only one of the many policies available to player two.


We take the typical problem in combinatorial optimisation---Travelling Salesman Problem (TSP)---which is widely used  for many real-world applications. For our Solvers we utilise a standard deep learning-based solver as our base solver. Our framework is solver agnostic, it can be applied to improve the generalization performance of any existing solvers. For the Data Generator, we develop a learning-to-attacking technique under the two-player framework by adding perturbations on top of uniformly generated data where the induced task instance can challenge the current Solver and make it perform poorly; as such, the generator can learn to  find the weaknesses of the Solver \footnote{See how easily the current TSP solvers can be exploited in  Appendix~\ref{app:attack effects}.}.

Overall, our contributions  are as follows:
\begin{itemize}
    \item We  study the generalization ability of TSP solvers from a game-theoretical perspective, and propose a two-player game framework to train effective Solvers which can incorporate, with minimal changes, any existing  deep learning-based Solver. 
    \item We propose a learning-to-attack method by adding \textit{learnable} perturbations on the data distribution of problem instance so that the solver can be exploited to perform poorly.
    \item We introduce a mixing-model by combining the population of Solvers so that we can make full use of the obtained solver population to attain the state-of-the-art results.
    \item We study the exploitability of obtained strategies during training to offer insights about how the solver population develop over time. Experimental results show that the obtained strategies under our framework are asymptotically approximating the Nash Equilibrium. 
\end{itemize}

\section{Related Work}
\textbf{Deep Learning for Combinatorial Optimization.} 
Deep learning or reinforcement learning based methods have achieved notable progress in various combinatorial optimization problems, such as Traveling Salesman Problem~\citep{lu2019learning, DBLP:conf/iclr/KoolHW19,kool2021deep}, Capacitated Vehicle Routing Problem~\citep{hottung2020learning,wu2021learning}, graph matching problems~\citep{yu2019learning, yu2021deep}. However, these works only focus on the data from uniform distribution, which ignore the generalization ability to unseen instances.

\textbf{Meta-Game Analysis.}
In meta-game analysis \citep{wellman2006methods,yang2020overview}, traditional solution concepts (e.g., Nash equilibrium) can be computed in a more scalable manner. PSRO~\citep{lanctot2017unified} generalises Double Oracle~\citep{mcmahan2003planning} by introducing RL to obtain an approximate best response.
In games with high degree of non-transitivity \citep{czarnecki2020real} such as Chess \citep{sanjaya2021measuring}, PSRO methods prove to be an efficient approach to prevent from learning strategic cycles.  $\text{PSRO}_{rN}$~\citep{balduzzi2019open} and Diverse-PSRO \citep{nieves2021modelling,DBLP:journals/corr/abs-2106-04958} incorporated diversity seeking into PSRO and Pipeline-PSRO~\citep{mcaleer2020pipeline} aims to improve training efficiency by training multiple best responses in parallel.

\section{Notations and Preliminaries}

\textbf{Normal Form Game (NFG)} - A tuple $(\Pi,U^{\Pi},n)$ where $n$ is the number of players, $\Pi=(\Pi_1,\Pi_2,...,\Pi_n)$ is the joint policy set and $U^{\Pi}=(U^{\Pi}_1,U^{\Pi}_2,...U^{\Pi}_n):\Pi\rightarrow\mathcal{R}^n$ is the utility matrix for each joint policy. A game is symmetric if all players have the same policy set ($\Pi_i=\Pi_j,i\ne j$) and same payoff structures, such that players are interchangeable. 

\textbf{Best Response} - The strategy which attains the best expected performance against a fixed opponent strategy. $\sigma_{i}^{*}=\text{br}(\Pi_{-i},\sigma_{-i})$ is the best response to $\sigma_{-i}$ if:
\begin{equation*}
\vspace{-1mm}
    U^{\Pi}_i(\sigma^{*}_i,\sigma_{-i})\ge U^{\Pi}_i(\sigma_i,\sigma_{-i} ), \forall i,\sigma_i\ne \sigma^{*}_i
\vspace{-1mm}
\end{equation*}

\textbf{Nash Equilibrium (NE)} - A strategy profile $\sigma^{*}=(\sigma^{*}_1,\sigma^{*}_2,...,\sigma^{*}_n)$ such that:
\begin{equation*}
U^{\Pi}_i(\sigma^{*}_i,\sigma^{*}_{-i})\ge U^{\Pi}_i(\sigma_i,\sigma^{*}_{-i} ), \forall i,\sigma_i\ne \sigma^{*}_i
\end{equation*}
Intuitively, no player has an incentive to deviate from their current strategy if all players are playing their respective Nash equilibrium strategy.

\textbf{Exploitability} - In a two-player zero-sum game, the average exploitability of a strategy profile $\sigma=(\sigma_1,\sigma_2)$ is defined as follows~\citep{davis2014using}:
\begin{equation}
\vspace{-1.5mm}
\label{exp}
    \text{exploitability}(\sigma) = \frac{1}{2}\big(U^{\Pi}_1(\text{br}(\sigma_{2}),\sigma_{2}\big)+U^{\Pi}_2\big(\sigma_1,\text{br}(\sigma_{1})\big)
    \vspace{-1.2mm}
\end{equation}

\textbf{Instance} - An individual sample of a combinatorial optimization problem. Hereafter, we denote an instance by $\mathcal{I}$ which comes from some distribution $\mathbf{P}_{\mathcal{I}}$. Specifically for TSP, instances represent a set of coordinates $\{(x_i,y_i)\in\mathcal{R}^n|i=1,2,...,n\}$ sampling repeatedly from some distribution.  

\textbf{Optimality gap} - Measures the quality of a Solver compared to an optimal Oracle. Given an Instance $\mathcal{I}$ and a Solver $S:\{\mathcal{I}\}\rightarrow\mathbf{R}$, the optimality gap is defined as:
\begin{equation}\label{og}
\vspace{-1mm}
    g(S,\mathcal{I}, \text{Oracle})=\frac{S(\mathcal{I})-\text{Oracle}(\mathcal{I})}{\text{Oracle}(\mathcal{I})}
    \vspace{-1mm}
\end{equation}
where Oracle$(\mathcal{I})$ gives the true optimal value of the Instance. Furthermore, the \textit{expected} optimality gap of an Instance distribution $\mathbf{P}_{\mathcal{I}}$ and an Oracle is defined as:
\begin{equation}\label{eog}
    G(S,\mathbf{P}_{\mathcal{I}},\text{Oracle})=\mathbf{E}_{\mathcal{I}\sim\mathbf{P}_{\mathcal{I}}}g(S,\mathcal{I},\text{Oracle}).
\end{equation}


\section{Our Method}\label{method}
In this section, we present our method for solving the Traveling Salesman Problem (TSP) at the meta-level.
Let there be two players in the meta-game, one is the Solver Selector (SS) and the other is the Data Generator (DG). $\Pi_{\text{SS}}=\{S_i|i=1,2,...\}$ is the policy set of candidate Solvers for the Solver Selector, and $\Pi_{\text{DG}}=\Pi_{N}\times\Pi_{C}=\{\mathbf{P}_{\mathcal{I},i}=(\mathbf{P}_{N,i},\mathbf{P}_{C,i})|i=1,2,...\}$ is the policy set for the Data Generator, where $\Pi_N$ is the policy set of the problem scale (i.e. the number of nodes that need to be generated) and $\Pi_C$ is the policy set used to generate two-dimensional coordinates. Therefore, an Instance distribution $\mathbf{P}_{\mathcal{I}}\in\Pi_\text{DG}$ comprises of two parts: $\mathbf{P}_{\mathcal{I}}=(\mathbf{P}_{N},\mathbf{P}_{C})$ where $\mathbf{P}_{N}$ is the distribution for the number of nodes $N$ contained in each Instance, and $\mathbf{P}_{C}$ is a two-dimensional distribution for coordinate positioning.


Formally, we have a two player zero-sum asymmetric NFG $(\Pi, \mathbf{U}^{\Pi}, 2)$ where $\Pi=(\Pi_{\text{SS}},\Pi_{\text{DG}})$, $\mathbf{U}^{\Pi}:\Pi\rightarrow\mathbf{R}^{|\Pi_{\text{SS}}|\times |\Pi_{\text{DG}}|}$, $\mathbf{U}^{{\Pi}_{\text{SS}}}(\pi)=G(\pi,\text{Oracle})$ is the expected optimality gap under the joint policy $\pi=(S,\mathbf{P}_{\mathcal{I}})\in\Pi$ as defined in Eq.~\ref{eog} and $\mathbf{U}^{{\Pi}_{\text{DG}}}(\pi)=-\mathbf{U}^{{\Pi}_{\text{SS}}}(\pi)$. Given $\mathbf{U}^{\Pi}$, we can determine a Nash Equilibrium $\sigma^{*}=(\sigma_{\text{SS}}^*,\sigma_{\text{DG}}^*)$ as the meta-strategy which satisfies: 
\begin{equation}
\vspace{-1mm}
    \min_{\sigma_{\text{SS}}\in\Delta(\Pi_{\text{SS}})}\max_{\sigma_{\text{DG}}\in\Delta(\Pi_{\text{DG}})}\mathrm{E}_{\pi\sim(\sigma_{\text{SS}},\sigma_{\text{DG}})}G\big(\pi,\text{Oracle}\big).
\vspace{-1mm}    
\end{equation}

We follow the PSRO framework as follows: at each iteration given the policy sets $\Pi_{\text{SS}}$ and $\Pi_{\text{DG}}$ and the meta-strategy $\sigma=(\sigma_{\text{SS}},\sigma_{\text{DG}})$ we train two Oracles:
\vspace{-2mm}
\begin{itemize}
    \item $S^{'}$ represents the Solver Selector, a best response to the Data Generator's meta strategy $\sigma_{\text{DG}}$.
    \item $\mathbf{P}_{\mathcal{I}}^{'}$ is the Data Generator which learns a data distribution where $\sigma_{\text{SS}}$ performs poorly.
\end{itemize}
\vspace{-2mm}
 Given these oracles, we update the joint policy set $\Pi^{'}=\Pi\cup(S^{'},\mathbf{P}_{\mathcal{I}}^{'})$ and the meta-game $\mathbf{U}^{\Pi^{'}}$ according to $\Pi^{'}$. This expansion of the joint policy set has a dual purpose in that it aids in finding the difficult to find Instance distributions whilst also improving the ability of the Solver Selector to solve said distributions. In line with our objective, this process leads to a population of powerful Solvers which have diverse abilities on various distributions, and how we successfully combine these individual Solvers is discussed in Sec.~\ref{employment of Solvers}.
The general algorithm framework can be seen in Alg.~\ref{alg:PSRO for CO}.

\begin{figure}[t!]
\vspace{-20pt}
\centering
\includegraphics[scale=.4]{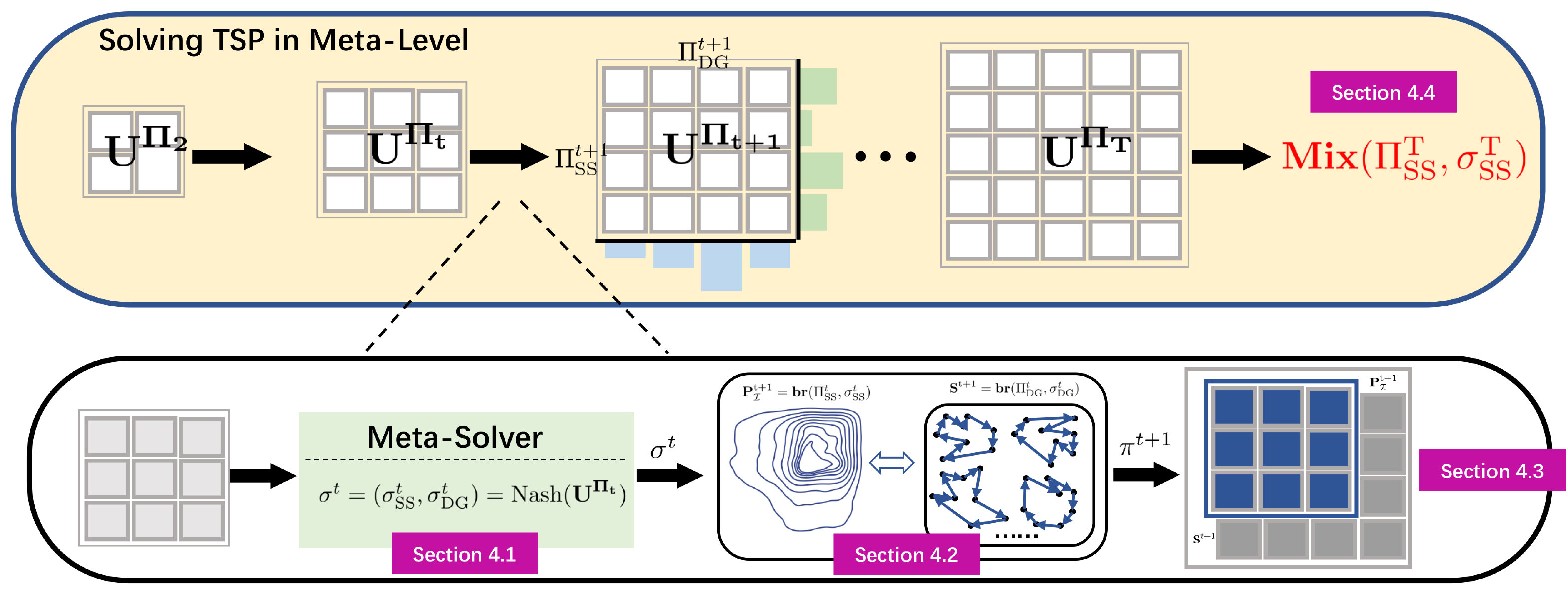}
\caption{Pipeline of solving combinatorial optimization problems in meta-level. At PSRO loop $t$, we first use meta-Solver to compute the meta-strategy $\sigma_t$ given the meta-table $\textbf{U}^{\Pi_t}$ and then training best response $(\textbf{S}^{t+1},\textbf{P}_{\mathcal{I}}^{t+1})$ based on current policy set and meta-strategy $(\Pi^{t},\sigma^{t})$. Finally we get a new meta-table $\textbf{U}^{\Pi_{t+1}}$ according to the new obtained policy and algorithm process iterates to the next loop.}
\label{fig:pipeline} 
\vspace{1mm}
\end{figure}

The formulation above leaves us with four algorithm components to address:  \textbf{1)} How to obtain the meta-strategy $\sigma$;    \textbf{2)} How to train Oracles for both players; \textbf{3)} How to evaluate the utilities $U^{\Pi}$; \textbf{4)} How to combine the the population of Solvers. 
In the following parts, we will describe the flow of the algorithm as visualised in Fig.~\ref{fig:pipeline}, where each section shown in purple represents the answer to the above four questions. 

\subsection{Meta-Strategy Solvers}

In this paper, we use the NE of the meta game as the meta-strategy solver as we believe for a two player zero-sum game the NE is sufficient, but various meta-strategy solvers can be used w.r.t the corresponding meta-game constructed by the specific combinatorial optimization problem.

\subsection{Oracle Training}
We now provide the higher-level details for training a best-response Oracle in the TSP setting, with more detailed derivations presented in Appendix~\ref{app:derive gradient}. Here we represent the trainable Solvers in $\Pi_{\text{SS}}$ as $S_{\theta}$ and the trainable Instance distributions in $\Pi_{\text{DG}}$ as $\mathbf{P}_{\mathcal{I},\gamma}=(\mathbf{P}_{N,\gamma_{N}},\mathbf{P}_{C,\gamma_{C}})$ where $\theta$ and $\gamma$ are the trainable parameters. $S_{\theta}$ can be any DL-based methods parameterized by ${\theta}$ so we can take derivation w.r.t. it to get the oracle for a given distribution.

\textbf{Solver Oracle.} Given the Data Generator meta-strategy $\sigma_{\text{DG}}$, the Oracle training objective for the Solver is:
\begin{equation}
\vspace{-1mm}
    \min_{\theta}L_{\text{SS}}(\theta)=\mathbf{E}_{\mathbf{P}_{\mathcal{I}}\sim\sigma_{\text{DG}}}G(S_\theta,\mathbf{P}_{\mathcal{I}},\text{Oracle}).
\end{equation}
The gradient of this objective is:
\begin{equation}\label{gradient of LSS}
\vspace{-2mm}
\begin{aligned}
&\nabla_{\theta}L_{\text{SS}}(\theta)
    &=\mathbf{E}_{\mathbf{P}_{\mathcal{I}}\sim\sigma_{\text{DG}}}\mathbf{E}_{N\sim\mathbf{P}_{N}}\mathbf{E}_{x_1,...,x_N\sim\prod_{i=1}^{N}\mathbf{P}_C}\frac{\nabla_{\theta}S_\theta(x_1,...,x_N)}{\text{Oracle}(x_1,...,x_N)}.
\end{aligned}
\vspace{-2mm}
\end{equation}

\textbf{Data Generator Oracle.} Given the Solver Selector meta-strategy $\sigma_{\text{SS}}$, the Oracle training objective for the Data Generator is:
\begin{equation}
\vspace{-2mm}
    \max_{\gamma}L_{\text{DG}}(\gamma)=\mathbf{E}_{S\sim\sigma_{\text{SS}}}G(S,\mathbf{P}_{\mathcal{I},\gamma},\text{Oracle}).
\end{equation}
To derive the gradient for the Data Generator Oracle, we first note the following: For $\Pi_N$, we fix the problem scale $\mathcal{N}=\{N_1,N_2,...\}$, and let $\mathbf{P}_{N,\gamma_N}\in\Pi_N$ be a parameterized discrete distribution over $\mathcal{N}$. Here we directly use a learnable probability vector $\gamma_{N}$ (implement by a softmax function) to represent $\mathbf{P}_{N,\gamma_N}$.

As the goal of the Data Generator is to find a suitable distribution that the current Solver finds difficult to solve, we design an approach that directly attacks the Solver by adding noise to given instances\footnote{Demonstrations of these attacks are shown in Appendix~\ref{app:attack effects}}. We achieve this by utilising an \textit{attacked} distribution where Instances sampled from a uniform distribution are perturbed by Gaussian noise. Formally, we start by sampling $\mathcal{I}\sim\text{U}(0,1)$, then we use a neural network parameterised attack generator $f_{\gamma_C}$ to generate the variance of a Gaussian distribution:
\begin{equation}\label{attack sigma}
\Sigma=f_{\gamma_C}(\mathcal{I})
\end{equation}
where the shape of $\Sigma$ is the same as $\mathcal{I}$, that is, if $\mathcal{I}$ contains $N$ two dimension coordinates then the variance matrix will be $\Sigma\in\mathrm{R}^{N\times 2}$, and we attack an Instance $\mathcal{I}$ additively via $\Tilde{\mathcal{I}}_{i,j} = \mathcal{I}_{i,j} + \epsilon$ where $\epsilon\sim\text{N}(0, \Sigma_{i,j})$. We denote the final attacked distribution by $\mathbf{P}_{C,\gamma_C}$, and our objective therefore is to find the optimal parameters $\gamma^{*}=(\gamma_C^{*},\gamma_N^{*})$.

The gradient w.r.t. $\gamma_C$ is:
\begin{equation}\label{gradient of attack1 TSP}
\vspace{-2mm}
\begin{aligned}
    \nabla_{\gamma_C}L_{\text{DG}}(\gamma)
    &=\mathbf{E}_{S\sim\sigma_{\text{SS}}}\mathbf{E}_{N\sim\mathbf{P}_{N,\gamma_N}}\mathbf{E}_{x_1,..,x_N\sim\prod_{i=1}^{N}\mathbf{P}_{C,\gamma_C}}\\ 
    & \qquad \qquad \Big[\nabla_{\gamma_C}\big(\sum_{i=1}^{N}\log\mathbf{P}_{C,\gamma_C}(x_i)\big)g\big(S,(x_1,...x_N),\text{Oracle}\big) \Big].
\end{aligned}
\end{equation}
Details relating to the extra computation required for the log-probability in Eq.~\ref{gradient of attack1 TSP} is left to Appendix~\ref{app: log prob compute}.

The gradient w.r.t. $\gamma_N$ is:
\begin{equation}\label{gradient of attack2 TSP}
\vspace{-1mm}
\begin{aligned}
    \nabla_{\gamma_N}L_{\text{DG}}(\gamma)
    &=\mathbf{E}_{S\sim\sigma_{\text{SS}}}\mathbf{E}_{N\sim\mathbf{P}_{N,\gamma_N}} \nabla_{\gamma_N}(\log\mathbf{P}_{N,\gamma_N}(N))\cdot\\ & \qquad \qquad  \quad \mathbf{E}_{x_1,..,x_N\sim\prod_{i=1}^{N}\mathbf{P}_{C,\gamma_C}}g\big(S,(x_1,...x_N\big),\text{Oracle}).
\end{aligned} 
\end{equation}
Overall, we have that the gradient of the Data Generator Oracle is:
 \begin{equation}\label{gradient of LDG}   
  \nabla_{\gamma}L_{\text{DG}}(\gamma)=
\left(               
  \begin{array}{c}
    \nabla_{\gamma_C}L_{\text{DG}}(\gamma)\\  
    \nabla_{\gamma_N}L_{\text{DG}}(\gamma)\\ 
  \end{array}
\right)             
\end{equation}

\textbf{Remark:} We omit the calling of 'Oracle' in Eq.~\ref{gradient of LSS} and Eq.~\ref{gradient of attack2 TSP} during implementation because the goals remain approximately same.

\subsection{Evaluation}
Given the joint policy set $\Pi$, we can compute the elements in matrix $U^{\Pi}$ by approximating the expected optimality gap defined in Eq.~\ref{eog}:
\begin{equation*}
    u_{S,\mathbf{P}_{\mathcal{I}}} = G(S,\mathbf{P}_{\mathcal{I}},\text{Oracle})=\mathbf{E}_{\mathcal{I}\sim\mathbf{P}_{\mathcal{I}}}g(S,\mathcal{I},\text{Oracle})\approx\frac{1}{M}\sum_{i=1}^{M}g(S,\mathcal{I}_i,\text{Oracle}).
    \vspace{-2mm}
\end{equation*}
where $S\in\Pi_{SS}, \mathbf{P}_{\mathcal{I}}\in\Pi_{DG}$.

\subsection{Combining the Solver population}\label{employment of Solvers}
At the end of training we are left with a \textit{diverse} population of Solvers designed for different distributions, which we suspect may be combined to generate a powerful general Solver. There already exist several works on how to mix policies, such as Q-Mixing ~\citep{smith2020learning, smith2021iterative}, however we instead choose to combine the Solvers based on the meta-strategy. As we use the Nash equilibrium as our meta-solver, we can guarantee that our combined Solver has a given \textit{conservative} level of performance under the assumption that these Instance distributions can be generated by the Data Generator's policy set. To some extent, we suggest that the conservative nature of the Nash equilibrium is itself in accordance with the meaning of generalization ability.
Additionally, in contrast to Q-mixing which only supports value-based methods, our mixing-method makes no prior assumptions on the type of RL algorithm in use, and can be used for both value-based and policy-based methods.

Formally, for value-based RL methods, we can weight the corresponding Q values to get the mixed Q-value of the combined model:
\begin{equation}\label{Q-mixing}
\vspace{-2mm}
\begin{aligned}
    Q_{\text{mix}}(s,a) =  \sum_{\pi\in\Pi_{\text{SS}}}\sigma^{*}_{\text{SS}}(\pi)Q_{\pi}(s,a)
\end{aligned}
\vspace{-1mm}
\end{equation}

For policy-based RL methods, we directly obtain the mixed policy probability by:
\begin{equation}\label{PG-mixing}
\vspace{-2mm}
\begin{aligned}
    \pi_{\text{mix}}(a|s) =\sum_{\pi\in\Pi_{\text{SS}}}\sigma^{*}_{\text{SS}}(\pi)\pi(a|s)
\end{aligned}
\end{equation}

\vspace{-10pt}
\section{Experiments}\label{experiment}
\begin{table}[!]
\vskip -10mm
\caption{Our model vs baselines. The gap \% is w.r.t. the best value across all methods.}
\label{results on generated data}
\centering
\footnotesize
\setlength{\tabcolsep}{0.35em}
\renewcommand{\arraystretch}{0.8}
\begin{tabular}{ll|rrr|rrr|rrr}
 & &  \multicolumn{3}{c|}{$n = 20$} & \multicolumn{3}{c|}{$n = 50$} & \multicolumn{3}{c}{$n = 100$} \\
 & Method &  \multicolumn{1}{c}{Obj.} & \multicolumn{1}{c}{Gap} & \multicolumn{1}{c|}{Time} & \multicolumn{1}{c}{Obj.} & \multicolumn{1}{c}{Gap} & \multicolumn{1}{c|}{Time} & \multicolumn{1}{c}{Obj.} & \multicolumn{1}{c}{Gap} & \multicolumn{1}{c}{Time} \\
\midrule
\midrule
\multirow{21}{*}{\rotatebox[origin=c]{90}{TSP}}
 &  Concorde  &  $3.43$ & $0.00 \%$ & (6s) & $4.99$ & $0.00 \%$ & (1m) & $6.20$ & $0.00 \%$ & (3m) \\
 &  LKH3  &  $3.43$ & $0.00 \%$ & (2s) & $4.99$ & $0.00 \%$ & (27s) & $6.20$ & $0.00 \%$ & (3m) \\
 &  Gurobi  &  $3.43$ & $0.00 \%$ & (1s) & $4.99$ & $0.00 \%$ & (19s) & $6.20$ & $0.00 \%$ & (4m) \\
\cmidrule{2-11}
 & AM(gr.) 
 &  $3.45$ & $0.58 \%$ & (2s)  & $5.12$ & $2.61 \%$ & (4s) & $6.71$ & $8.23 \%$ & (8s) \\
 &  LIH(T=1000) 
 &  $3.69$ & $7.72 \%$ & (16s) & $5.07$ & $1.72 \%$ & (33s) & $6.72$ & $8.48 \%$ & (63s) \\
 &  \textbf{LIH(FS)(T=1000)}  &  $\mathbf{3.43}$ & $\mathbf{0.12 \%}$ & (18s) & $\mathbf{5.06}$ & $\mathbf{1.68 \%}$ & (34s) & $\mathbf{6.47}$ & $\mathbf{4.43 \%}$ & (67s) \\
 &  \textbf{LIH(FT)(T=1000)}  &  $\mathbf{3.43}$ & $\mathbf{0.04 \%}$ & (50s) & $\mathbf{5.07}$ & $\mathbf{1.70} \%$ & (64s) & $\mathbf{6.45}$ & $\mathbf{4.00 \%}$ & (2m) \\
\cmidrule{2-11}
   &  AM(sampling) 
   &  $3.43$ & $0.11 \%$ &(6s)& $5.03$ & $0.95 \%$ & (29s) & $7.22 $ & $16.45 \%$ & (2m)  \\
    &  LIH(T=3000) 
    &  $3.62$ & $5.54 \%$ & (46s) & $5.03$ & $0.92 \%$ & (95s) & $6.58$ & $6.24 \%$ & (3m) \\
    &  Att-GCRN+MCTS 
    &  $3.43$ & $\textbf{0.00\%}$ & ( - ) & $5.09$ & $2.16 \%$ & ( - ) & $6.65$ & $7.32 \%$ & ( - ) \\
    &  DPDP(bs=10K) 
    &  $3.43$ & $\textbf{0.00\%}$ & (5s) & $5.03$ & $1.00 \%$ & (3m) & $6.86$ & $10.66 \%$ & (12m) \\
 &  \textbf{LIH(FS)(T=3000)}  &  $\mathbf{3.43}$ & $0.04 \%$ & (54s) & $\mathbf{5.03}$ & $\mathbf{0.88 \%}$ & (100s) & $\mathbf{6.37}$ & $\mathbf{2.77 \%}$ & (3m) \\
 &  \textbf{LIH(FT)(T=3000)}  &  $\mathbf{3.43}$ & $\mathbf{0.00 \%}$ & (2m) & $\mathbf{5.03}$ & $\mathbf{0.89} \%$ & (3m) & $\mathbf{6.36}$ & $\mathbf{2.50 \%}$ & (6m) \\
\midrule
\midrule
\end{tabular}
\vspace{-2mm}
\end{table}

In the following section, we present our results on TSP Instances of size $n=20,50,100$. In contrast to previous work which fail to show generalization ability due to training and testing on the same distribution (uniform), we demonstrate performance on distributions that are unseen during training. We detail the basic settings below, with specific training settings listed in Appendix~\ref{app:exp settings}.

\subsection{Experimental Settings}
\textbf{Data normalization.} We only consider Instances within $[0,1]\times[0,1]$ which are normalised via min-max normalization:
\begin{equation}\label{normalization}
    \mathcal{I}_{norm} = \text{Norm}(\mathcal{I})= \frac{\mathcal{I}-\min\{\mathcal{I}\}}{\max\{\mathcal{I}\}-\min\{\mathcal{I}\}}
\end{equation}
where $\min\{\mathcal{I}\}$ is the minimum scalar coordinate value in the TSP Instance and  $\max\{\mathcal{I}\}$ is the max coordinate value. 

\textbf{Data generation.} We generate data by randomly sampling $x\in\mathrm{R}^2$ from the unit square, and sampling $y\in\mathrm{R}^2$ from $\text{N}(\textbf{0},\Sigma)$ where $\Sigma\in\mathrm{R}^{2\times2}$ is a diagonal matrix whose elements are sampled from $[0,\lambda]$ and $\lambda\sim\text{U}(0,1)$. Next, a two-dimensional coordinate is generated by $z=x+y$, and we can get any scale $n$ of TSP by performing this sampling $n$ times. We sample 1000 normalised (via Eq.~\ref{normalization}) Instances which makes up 10 groups of data generated by different $\lambda$ values, and report the relevant results on the generated datasets in Table~\ref{results on generated data}.

\textbf{Baselines.} In this work, our base RL model comes from \citep{wu2021learning} which we denote as \textbf{LIH}.
We compare our model with Gurobi~\citep{gurobi}, Concorde, LKH3~\citep{helsgaun2017extension} and the following deep learning-based methods: AM~\citep{DBLP:conf/iclr/KoolHW19},~\citep{wu2021learning}, Att-GCRN+MCTS~\citep{fu2020generalize}~\footnote{We don't report the time because the implementation is using C programming which is different from others} and DPDP~\citep{kool2021deep} on generated data. On the real world instances from TSPLib~\citep{reinelt1991tsplib}, we compare the results with known best results, Or-tools~\citep{ortools}, AM~\citep{DBLP:conf/iclr/KoolHW19} and LIH~\citep{wu2021learning}. All experiments are trained and executed with one single GPU (RTX3090) and CPU (i9-10900KF).
\begin{table}[!]
\vspace{-2mm}
\renewcommand{\arraystretch}{.99}
\centering 
\caption{Results on TSPlib Instances. The underlined  and bold figures mean achieving the best results among all methods (including OR-Tools) and all deep learning-based methods respectively.} 
\vspace{-1mm}
\begin{threeparttable}\resizebox{12cm}{!}{ 
\begin{tabular}{l|cc|ccccc} 
\midrule 
\multirow{2}{*}{\rotatebox[origin=c]{0}{\textbf{\makecell{Instance}}}} & \multirow{2}{*}{\rotatebox[origin=c]{0}{\makecell{Opt.}}} &
\multirow{2}{*}{\rotatebox[origin=c]{0}{\makecell{OR-Tools}}} & AM  & AM  & LIH & \textbf{LIH(FS)} & \textbf{LIH(FT)}  \\ 
 &   &  & $ (N$=1,280) & ($N$=5,000) & ($T$=3,000) & ($T$=3,000)& ($T$=3,000)\\
\midrule 
    eil51 & 426   & 436   & 435   & 434   & 438   & \textbf{\underline{429} } & \textbf{\underline{429} } \\
    pr124 & 59,030  & 62,519  & 62,750  & 61,996  & 66,010  & \textbf{\underline{61,645} } & \textbf{\underline{61,645}} \\
    rd100 & 7,910  & 8,189  & 8,180  & 8,048  & 7,915  & \textbf{\underline{8,036}} & 8,160  \\
    pr76  & 108,159  & 111,104  & 111,598  & 111,924  & 109,668  & 109,418  & \textbf{\underline{108,495}} \\
    kroB150 & 26,130  & 27,572  & 28,894  & 28,864  & 31,407  & \textbf{\underline{27,418}} & \textbf{\underline{27,418 }} \\
    u159  & 42,080  & 45,778  & 45,394  & 44,581  & 51,327  & \textbf{\underline{43,376}} & \textbf{\underline{43,376}} \\
    berlin52 & 7,542  & 7,945  & 9,759  & 9,831  & 8,020  & 7,653  & \textbf{\underline{7,544}} \\
    eil101 & 629   & 664   & 656   & 656   & 658   & \textbf{\underline{642}} & 656  \\
    kroC100 & 20,749  & 21,583  & 22,586  & 22,896  & 25,343  & \textbf{\underline{21,079}} & 21,255  \\
    eil76 & 538   & 561   & 558   & 557   & 575   & \textbf{\underline{548}} & \textbf{\underline{548}} \\
    kroB100 & 22,141  & 23,006  & 24,340  & 23,987  & 26,563  & \textbf{\underline{22,855}} & 23,677  \\
    kroE100 & 22,068  & 22,598  & 22,895  & 22,716  & 26,903  & \textbf{\underline{22,532}} & 22,898  \\
    bier127 & 118,282  & 122,733  & 130,513  & 128,150  & 142,707  & \textbf{\underline{127,520}} & \textbf{\underline{127,520}} \\
\midrule 
\midrule 
Avg. Gap ($\%$)  & 0  &3.46 &42.96 & 36.86& 17.12&\textbf{5.13}& 5.49\\ 
\bottomrule 
\end{tabular}}
\end{threeparttable}
\label{results on real problem} 
\vspace{-4mm}
\end{table}
\subsection{Results}
\textbf{Results on Generated Data.} 
We follow two different training paradigms based on LIH: training from scratch \textbf{LIH(FS)} and fine-tuning \textbf{LIH(FT)}, with specific settings available in Appendix~\ref{app:exp settings}.
We first evaluate how \textbf{LIH(FS)} and \textbf{LIH(FT)} perform on the unseen test data. In Table~\ref{results on generated data}, we can see that the performance of deep-learning based methods trained on a uniform distribution degrades when dealing with Instances from an unseen distribution. On the other hand, our model obtained from the PSRO framework performs well out-of-distribution, and achieves state-of-the-art results among the deep-learning methods. Notably, we do not tune any hyperparameters in the original RL Solver, which suggests these improvements are based solely on the different training paradigm introduced by our method. However, due to the use of a mixing-strategy, the time consumed grows linearly compared with the base Solver due to the extra feed-forward computation, which can be seen as a trade-off between solution quality and runtime.

\textbf{Results on Real-World Problems.} We also test our \textbf{LIH} variants on real TSP problems from TSPLib~\citep{reinelt1991tsplib}, and report these results in Table~\ref{results on real problem}. We maintain the settings from \citep{wu2021learning} with T$=3000$ representing the numbers of calling improvement heuristics, and the results reported  for OR-Tools~\citep{ortools} and \textbf{LIH} are directly taken from \citep{wu2021learning}. 

From the results in Table~\ref{results on real problem}, we show that our model has the best performance among deep-learning based methods in the majority of Instances from TSPLib~\cite{reinelt1991tsplib}, and for the subsection: eil51, pr124, rd100, pr76, kroB150, u159, eil101, kroC100, eil76, kroB100, kroE100, bier127, our method is able to outperform OR-Tools.

\section{Discussions}
In this section, we provide more insightful analysis into properties of the meta-games and the population Solvers.
\subsection{Meta game analysis}\label{MGA}
\begin{figure}[t!]
 \vspace{-10mm}
\centering
\subfigure[Exploitability on TSP20]{
\label{fig:exp20}
\includegraphics[scale=.25]{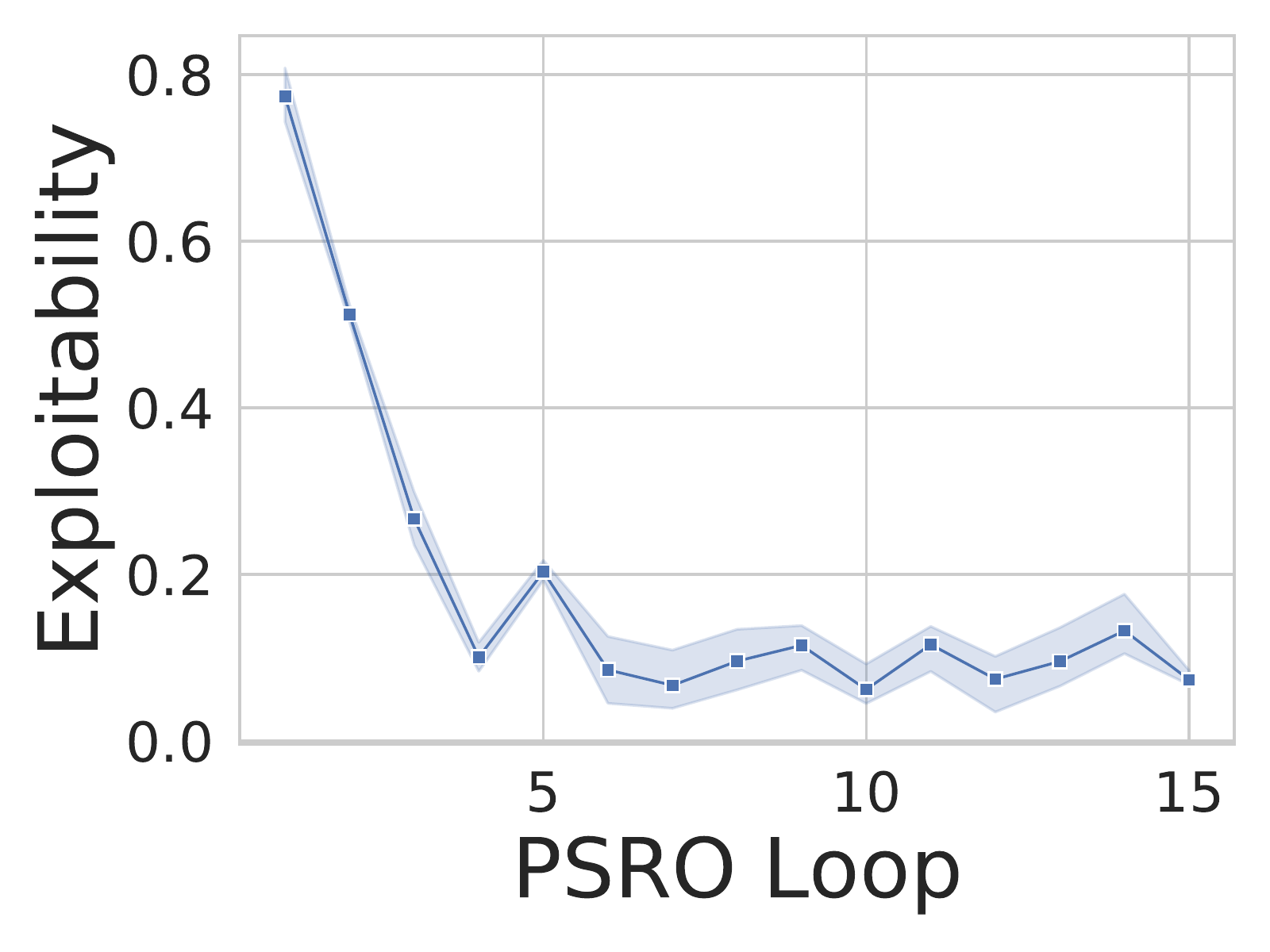}}
\hspace{.1in}
\subfigure[Exploitability on TSP50]{
\label{fig:exp50}
\includegraphics[scale=.25]{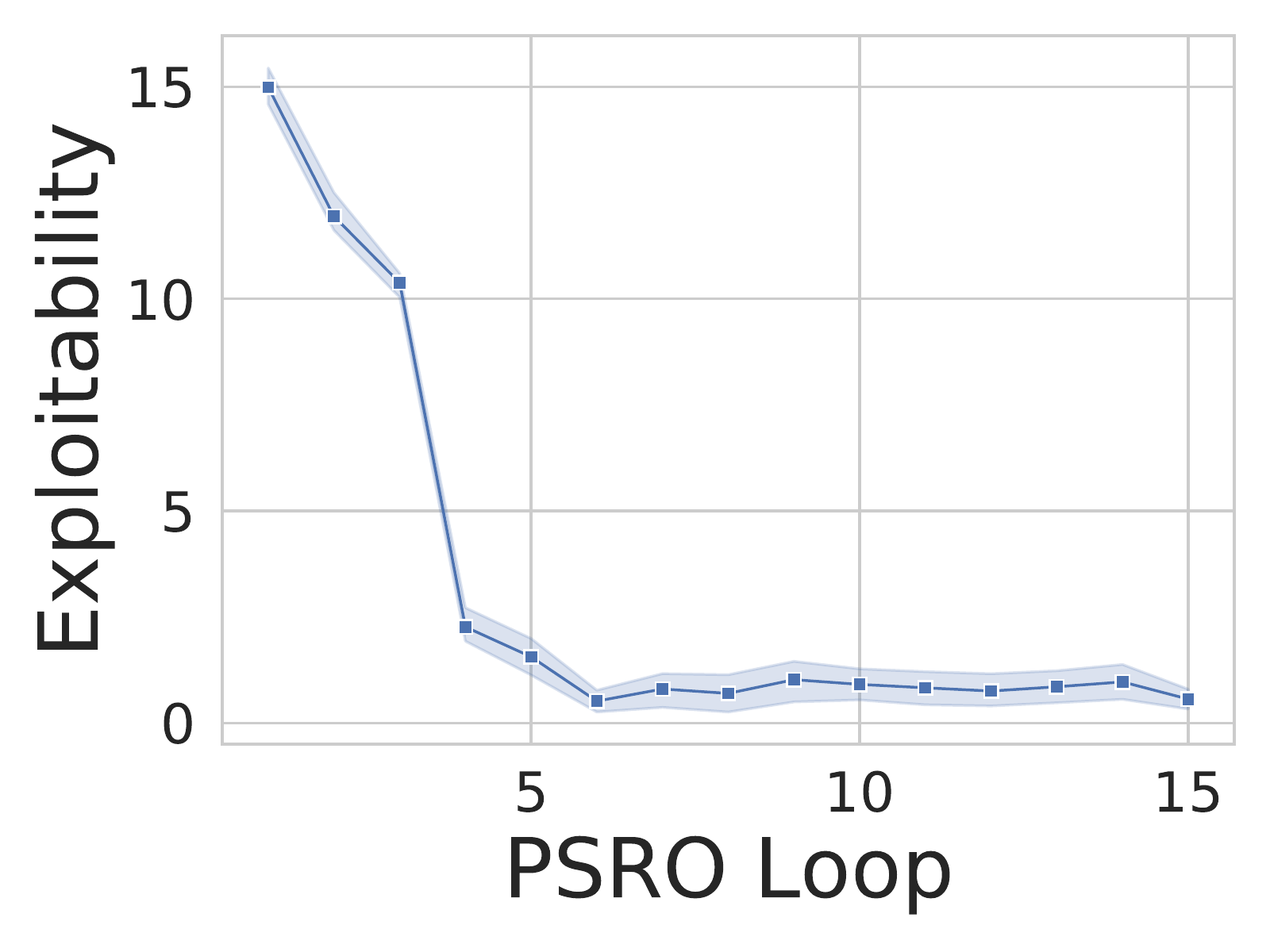}}
\hspace{.1in}
\subfigure[Exploitability on TSP100]{
\label{fig:exp100}
\vspace{-25pt}
\includegraphics[scale=.25]{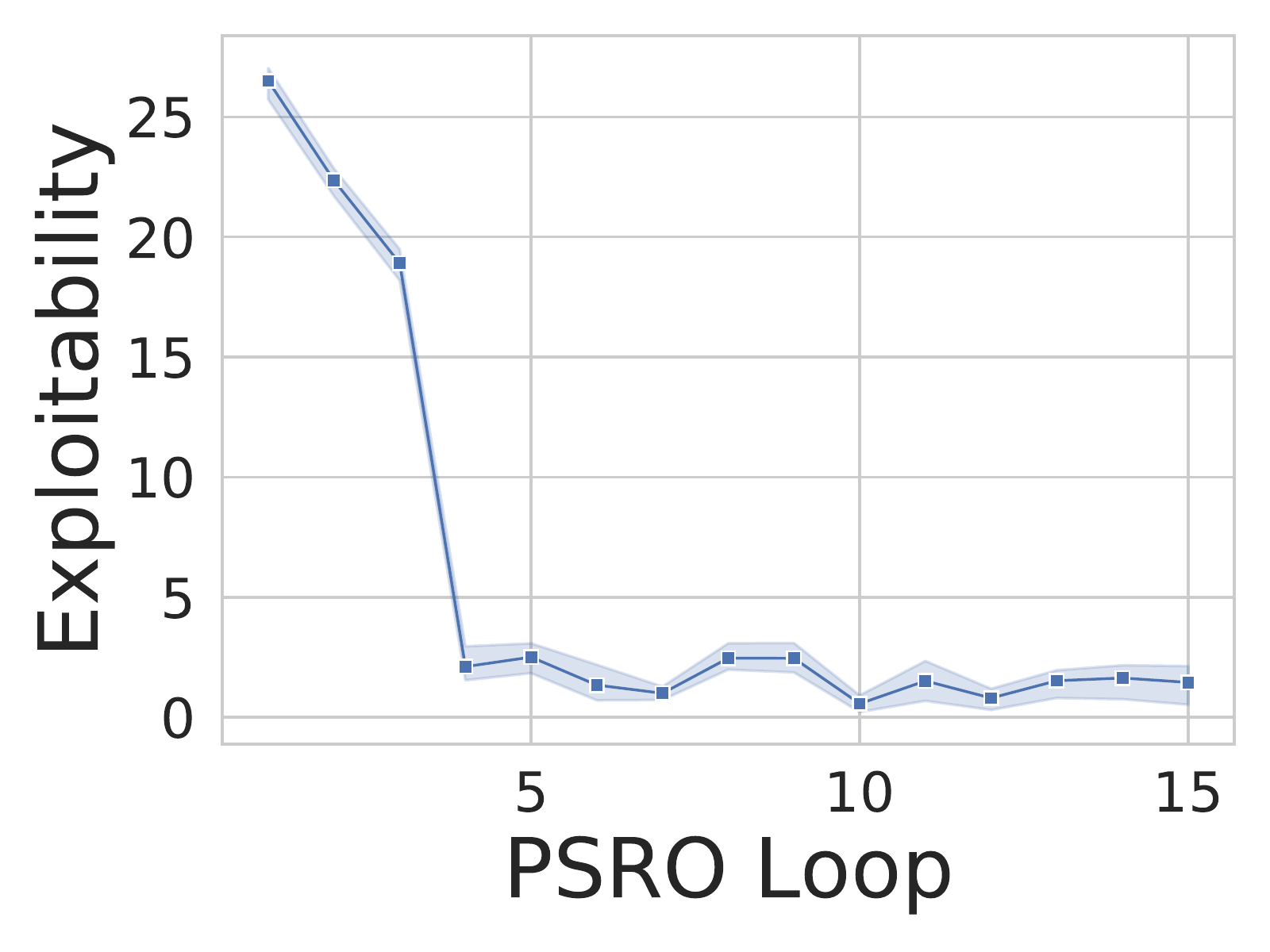}}
\hspace{.1in}
\subfigure[Opt. gap of combining Solvers trained on TSP20]{
\label{fig:abl_psro20}
\includegraphics[scale=.24]{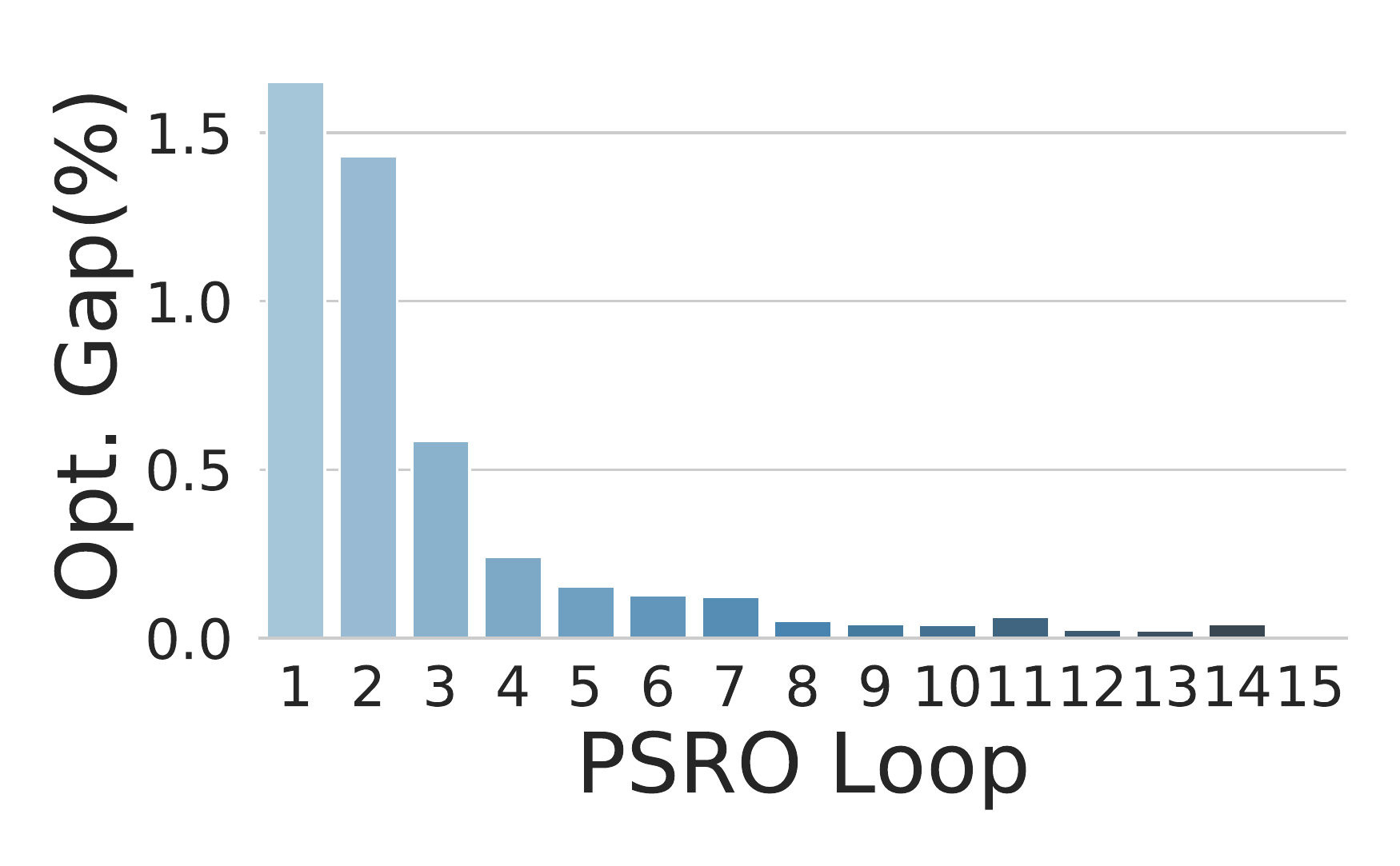}}
\hspace{0.1in}
\subfigure[Opt. gap of combining Solvers trained on TSP50]{
\label{fig:abl_psro50}
\includegraphics[scale=.24]{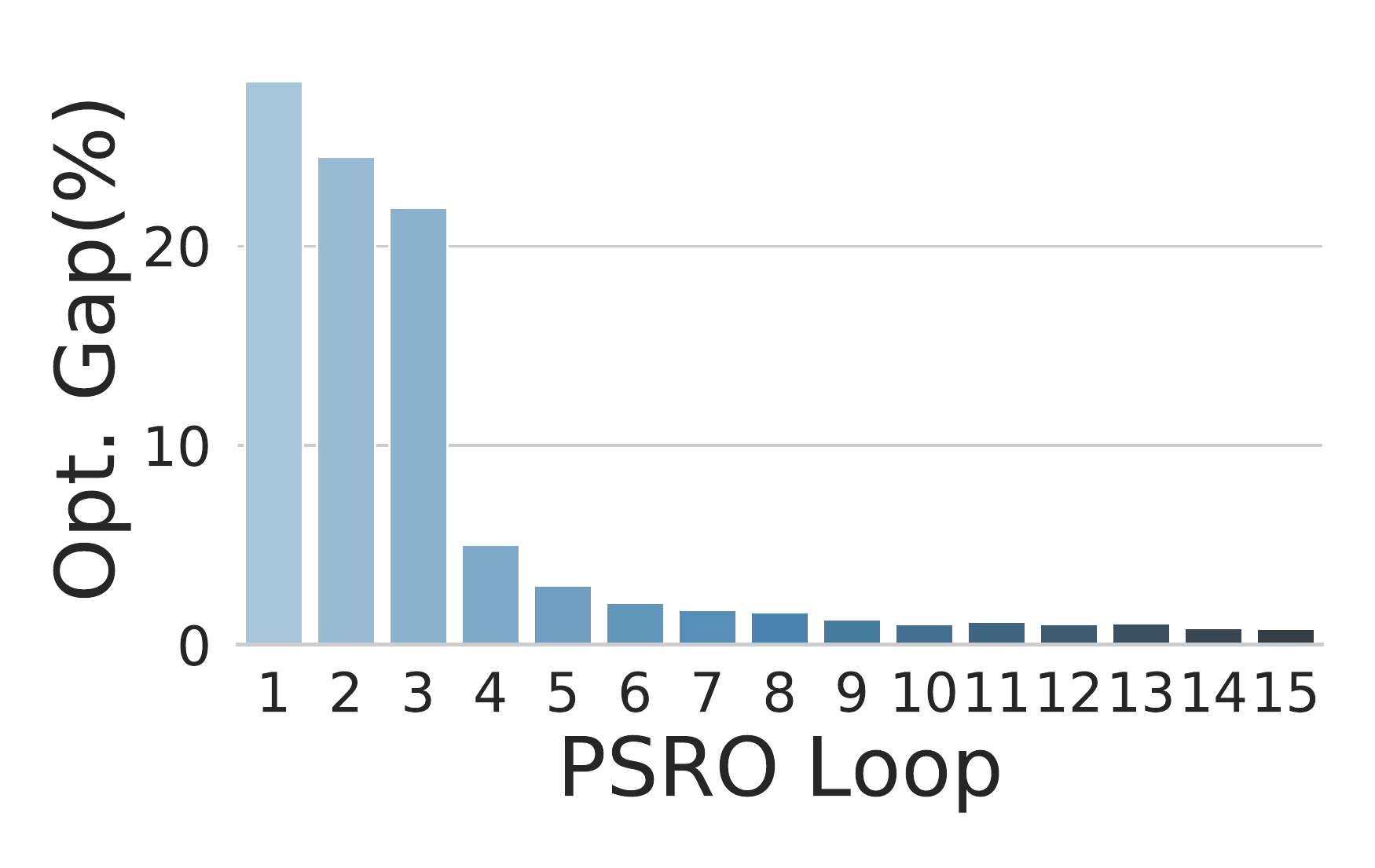}}
\hspace{0.1in}
\subfigure[Opt. gap of combining Solvers trained on TSP100]{
\label{fig:abl_psro100}
\includegraphics[scale=.24]{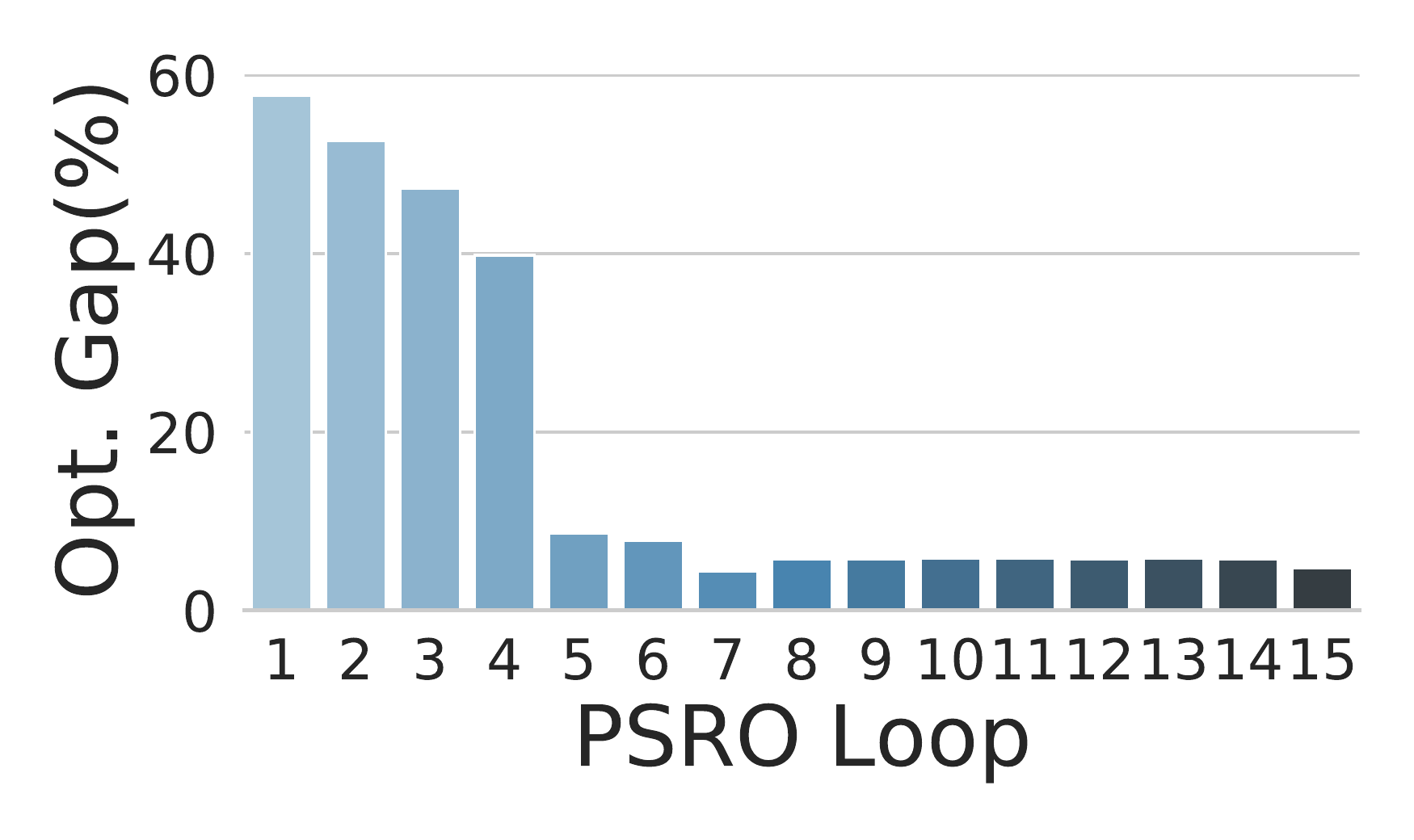}}
\vspace{-5mm}
\caption{Exploitability and performance of our model as the PSRO training goes on}
\label{fig:exp_psro} 
\end{figure}

To demonstrate the Game-Theoretic rationality of our method, we provide an analysis of the exploitability (Eq.~\ref{exp}) of our learned populations. Specifically, we train for 15 PSRO iterations on TSP20, TSP50, TSP100 and compute the exploitability of the obtained meta-strategy at each PSRO iteration. Results are shown in Fig.~\ref{fig:exp_psro}, with the visible decrease in exploitability demonstrating the validity of both our framework and the usage of the Nash equilibrium as the meta-solver. In particular, as training goes on, we generate strategies which approximately monotonically improve towards the Nash Equilibrium. 

\subsection{Usage of a population of Solvers}

In this section we discuss how restricting the number of Solvers available to be mixed-over, and how the choice of mixing weights may impact the final results. All results are obtained by \textbf{LIH(FS)} (T=1000), and we use Instances generated in the same manner as Section~\ref{experiment}. 

\begin{figure}[t!]
 \vspace{-5mm}
\centering
\subfigure{
\label{fig:abla_wn20}
\includegraphics[scale=.23]{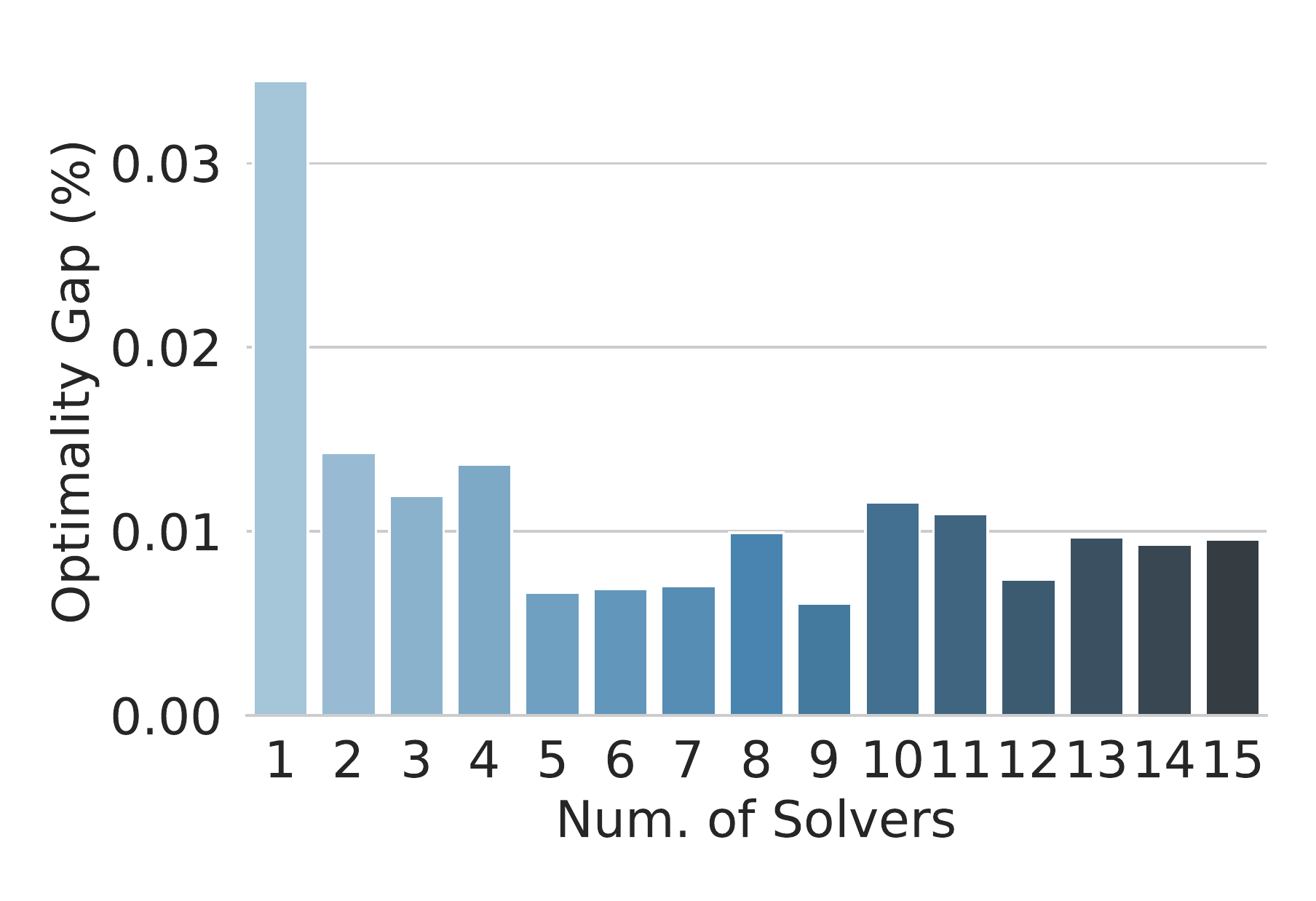}}
\subfigure{
\label{fig:abla_wn50}
\includegraphics[scale=.23]{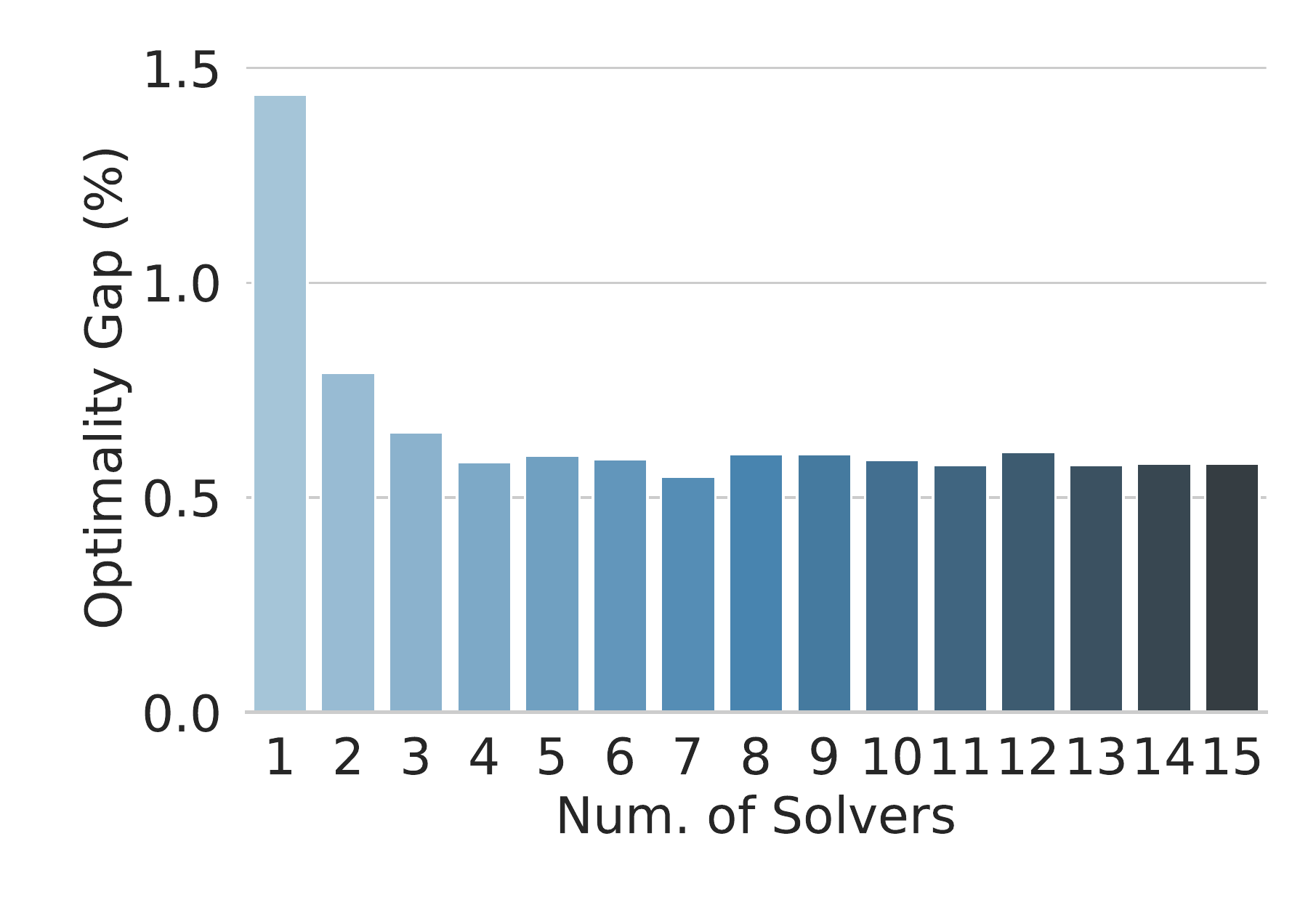}}
\subfigure{
\label{fig:abla_wn100}
\includegraphics[scale=.23]{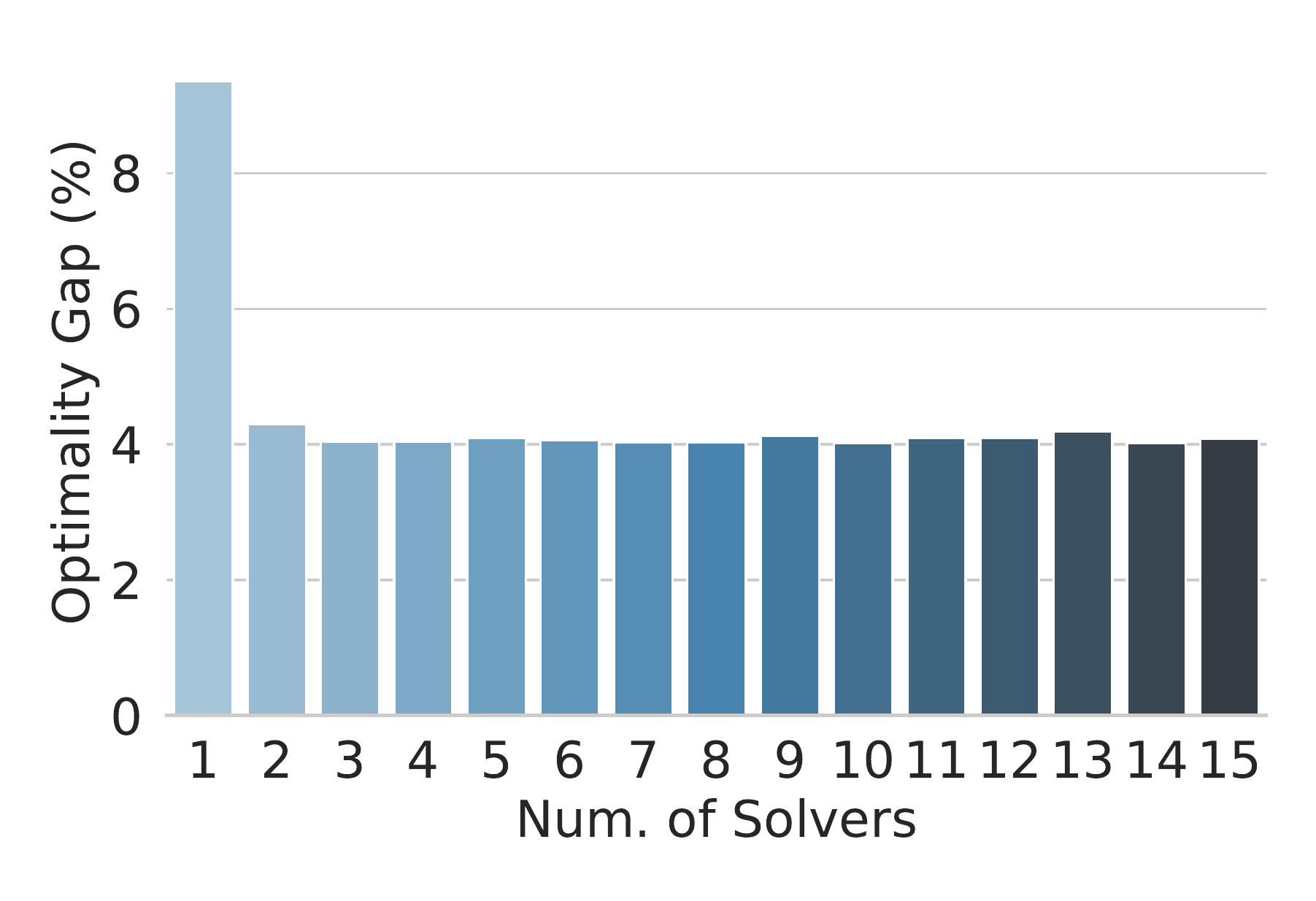}}
\vspace{-5pt}
\caption{Optimality gap of mixing-Solver with different combined numbers on TSP20, 50, 100 (from left to right).}
\label{fig:abla_num} 
\vspace{-3mm}
\end{figure}

We first investigate the impact of the number of Solvers used in the combined-model. The Solvers obtained by PSRO are ranked according to their corresponding density in the meta-strategy, and we combine the top-k Solvers showing the performance in Fig.~\ref{fig:abla_num}. Results show that our mixing method improves the performance significantly over the single most powerful Solver.
In practice, it's preferable to use fewer Solvers in the combined-Solver as this reduces the amount of resources required. In this paper, we set the the probability threshold of 0.99 to choose solvers (1-3 solvers usually) to balance the performance and computational cost.

\begin{figure}[t!]
\centering
\subfigure{
\label{fig:abla_wn20}
\includegraphics[scale=.1]{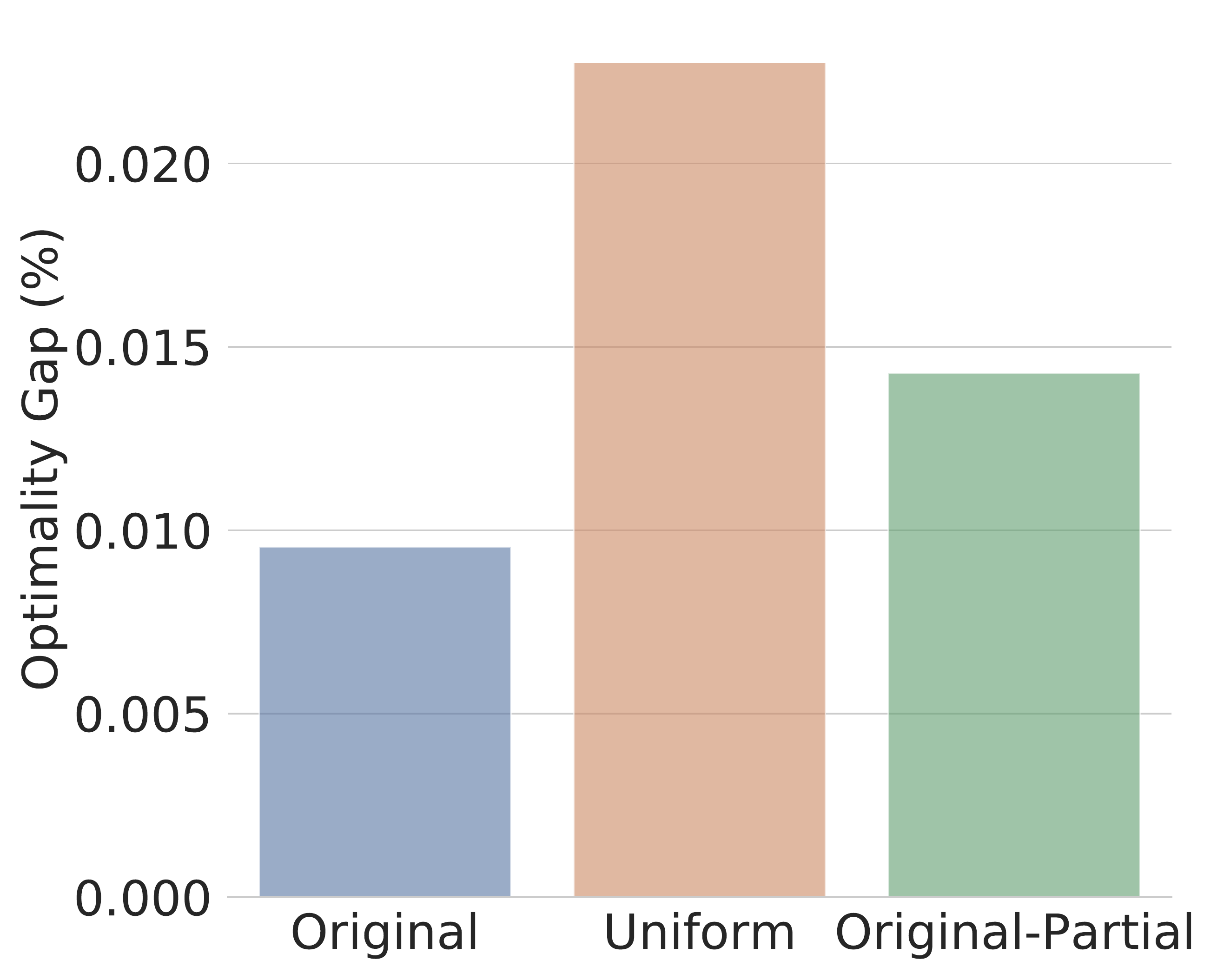}}
\hspace{.31in}
\subfigure{
\label{fig:abla_wn50}
\includegraphics[scale=.1]{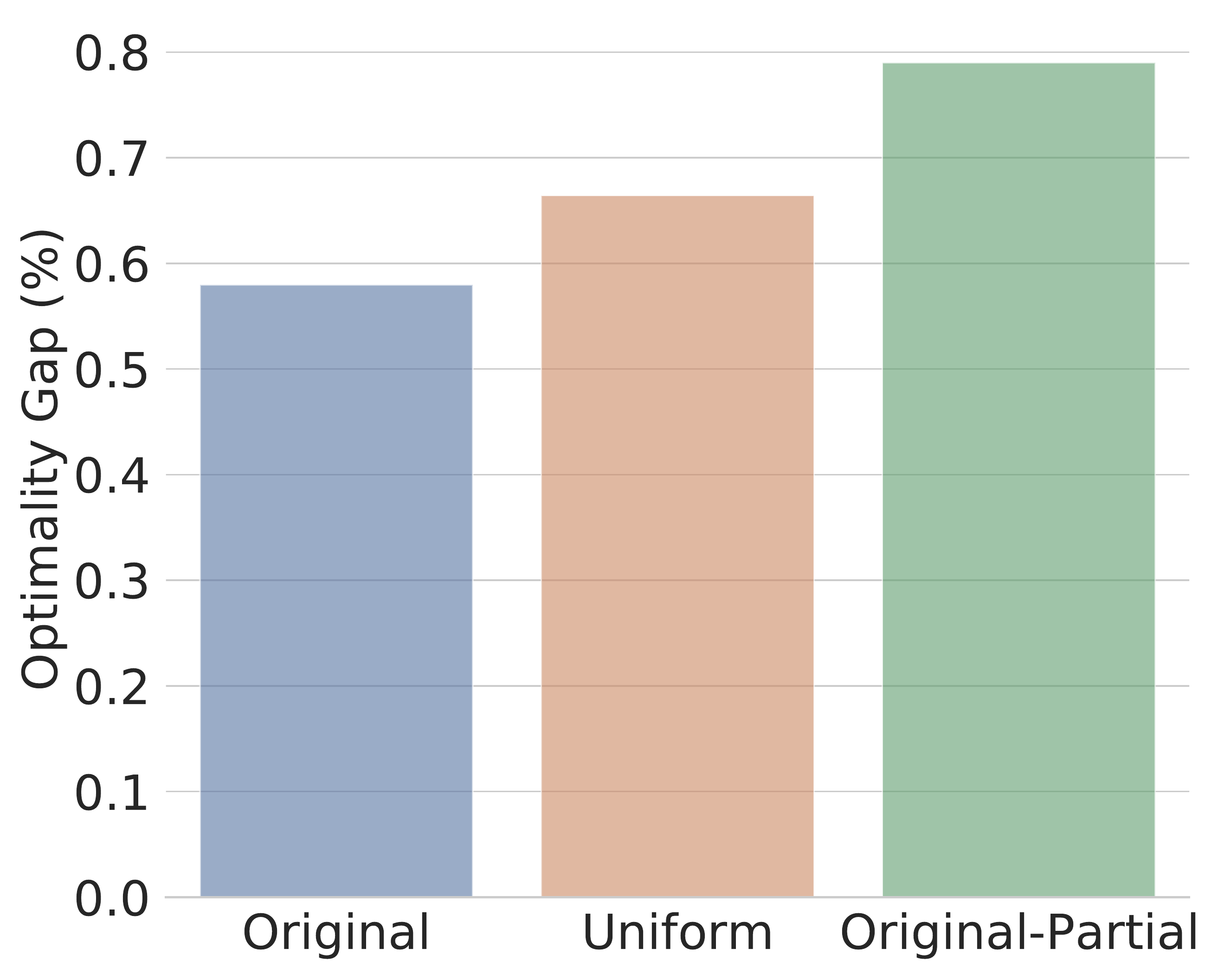}}
\hspace{.31in}
\subfigure{
\label{fig:abla_wn100}
\includegraphics[scale=.1]{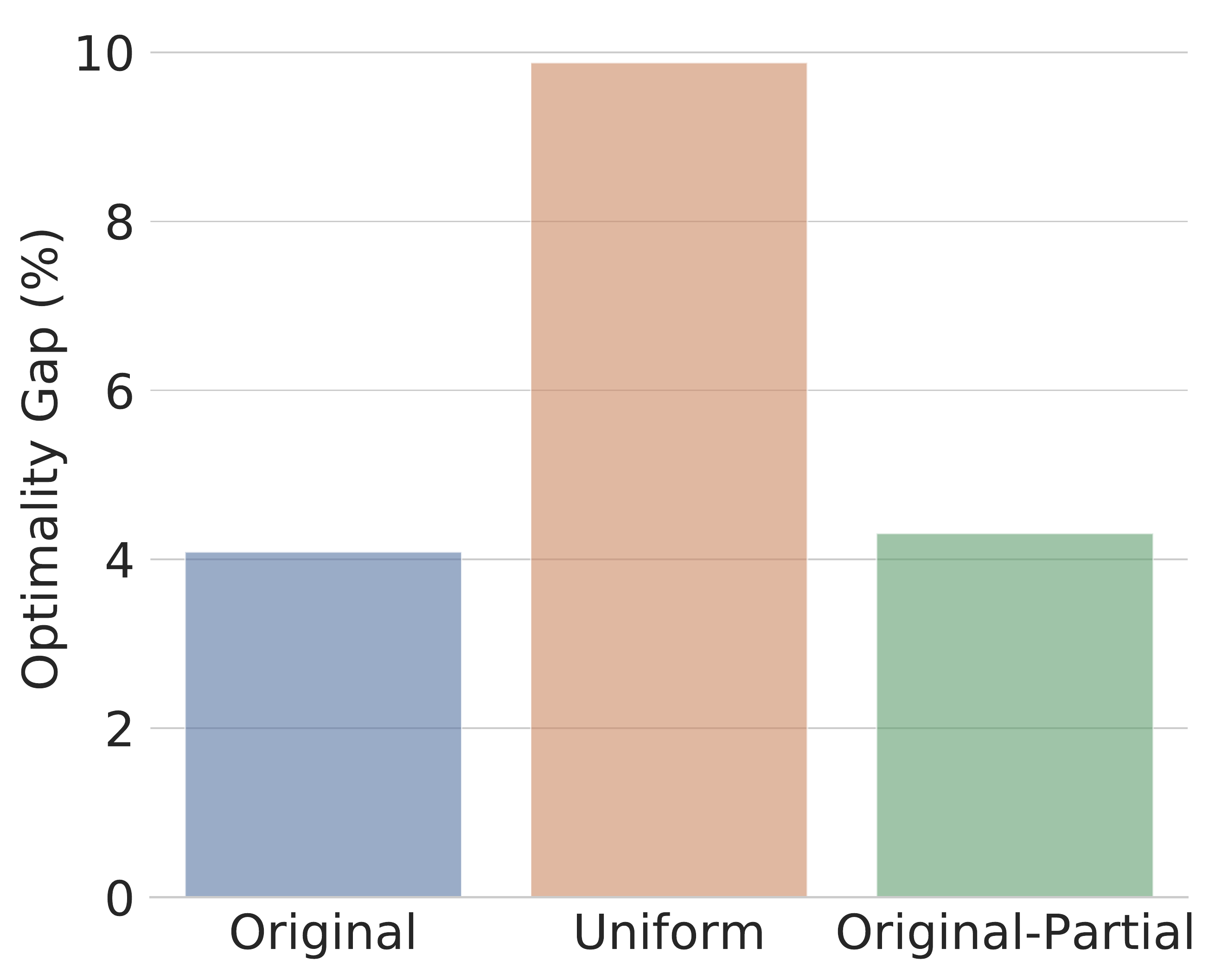}}
\vspace{-5mm}
\caption{Optimality gap of mixing-Solver with different combined numbers and weights on TSP20, 50, 100 (from left to right)}
\label{fig:abla_wn}
\end{figure}

Furthermore, we investigate the impacts of mixing-weight. Here we consider three scenarios: 
\begin{itemize}
\vspace{-2mm}
    \item Original: the whole population of Solvers combined by the Nash equilibrium meta-strategy.
    \vspace{-1mm}
    \item Uniform: the whole population of Solvers combined by a uniform meta-strategy.
        \vspace{-1mm}
    \item Original-Partial: the two most powerful Solvers as judged by the meta-strategy (i.e. they have the largest amount of density in the meta-strategy), combined by their respective normalized meta-strategy probabilities.
    \vspace{-5mm}
\end{itemize}

Fig.~\ref{fig:abla_wn} shows the comparisons between the different cases listed above. We can see the 'Original' setting achieves the best results, which shows the theoretical stability of the Nash equilibrium meta-solver as described in Section.~\ref{employment of Solvers}. In the case of the 'Uniform' setting, the performance degenerates as $n$ increases, which we suspect is due to the equal importance given to all Solvers, even if some of them are particularly weak. For the 'Original-Partial' setting, the use of only 2 Solvers violates our original framework, specifically leading to a reduction in Solver diversity and a poor ability to deal with unseen problems. 


\section{Conclusions}
Building on the framework of PSRO, in this paper, we  propose the first  game-theoretic solution to  improving the generalization ability for any neural TSP Solvers.  
On both randomly-generated and real-world TSP instances, we show that the Solvers trained under our two-player framework demonstrate the state-of-the-art generalization performance when compared to a series of strong TSP solution baselines.  
In principle, our proposed two-player game enables to improve the generalization of the Solver population by decreasing its exploitability against an adaptive data Generator, which gets increasingly stronger during training. 
In  future, we will apply such a game-theoretic  framework on other types of combinatorial optimization problems and explore the potentials on fine-tune techniques on this framework.

\section*{Acknowledgement}
This paper is supported by National Key R\&D Program of China (2021YFA1000403), the National Natural Science Foundation of China (Nos. 11991022), the Strategic Priority Research Program of Chinese Academy of Sciences (Grant No. XDA27000000) and the Fundamental Research Funds for the Central Universities.

\appendix
\section{Appendix}
\subsection{Oracle Training}\label{app:derive gradient}
We need train two oracles: $S^{'}$ and $\mathbf{P}^{'}_{\mathcal{I}}$ as a new policy to be added to the corresponding policy set. Here we will provide a specific derivation for training the oracle in our combinatorial optimization problems setting. 

Taken the formula from Eq.~\ref{gradient of LSS}, the gradient is apparent to get:
\begin{equation}\label{derive LSS}
\begin{aligned}
    \nabla_{\theta}L_{\text{SS}}(\theta)
    &=\nabla_{\theta}\mathbf{E}_{\mathbf{P}_{\mathcal{I}}\sim\sigma_{\text{DG}}}\mathbf{E}_{{\mathcal{I}}\sim\mathbf{P}_{\mathcal{I}}}g(S_\theta,{\mathcal{I}},\text{Oracle})\\
    &=\mathbf{E}_{\mathbf{P}_{\mathcal{I}}\sim\sigma_{\text{DG}}}\mathbf{E}_{{\mathcal{I}}\sim\mathbf{P}_{\mathcal{I}}}\nabla_{\theta}g(S_\theta,{\mathcal{I}},\text{Oracle})\\
    &=\mathbf{E}_{\mathbf{P}_{\mathcal{I}}\sim\sigma_{\text{DG}}}\mathbf{E}_{{\mathcal{I}}\sim\mathbf{P}_{\mathcal{I}}}\frac{\nabla_{\theta}S_\theta(\mathcal{I})}{\text{Oracle}(\mathcal{I})}\\
    &=\mathbf{E}_{\mathbf{P}_{\mathcal{I}}\sim\sigma_{\text{DG}}}\mathbf{E}_{N\sim\mathbf{P}_{N}}\mathbf{E}_{x_1,...,x_N\sim\prod_{i=1}^{N}\mathbf{P}_C}\frac{\nabla_{\theta}S_\theta(x_1,...,x_N)}{\text{Oracle}(x_1,...,x_N)}.
\end{aligned}
\end{equation}
Also for Eq.~\ref{gradient of LDG}, the computation of this gradient is:
\begin{equation}\label{derive LDG}
\begin{aligned}
    \nabla_{\gamma}L_{\text{DG}}(\gamma)&=\mathbf{E}_{S\sim\sigma_{\text{SS}}}\nabla_{\gamma}\mathbf{E}_{{\mathcal{I}}\sim\mathbf{P}_{\mathcal{I},\gamma}}g(S,{\mathcal{I}},\text{Oracle})\\
    &=\mathbf{E}_{S\sim\sigma_{\text{SS}}}\int_{\mathcal{I}}\nabla_{\gamma}\mathbf{P}_{\mathcal{I},\gamma}(\mathcal{I})g(S,{\mathcal{I}},\text{Oracle})\mathbf{d}\mathcal{I}\\
    &=\mathbf{E}_{S\sim\sigma_{\text{SS}}}\int_{\mathcal{I}}\mathbf{P}_{\mathcal{I},\gamma}(\mathcal{I})\frac{\nabla_{\gamma}\mathbf{P}_{\mathcal{I},\gamma}(\mathcal{I})}{\mathbf{P}_{\mathcal{I},\gamma}(\mathcal{I})}
    g(S,{\mathcal{I}},\text{Oracle})\mathbf{d}\mathcal{I}\\
    &=\mathbf{E}_{S\sim\sigma_{\text{SS}}}\mathbf{E}_{{\mathcal{I}}\sim\mathbf{P}_{\mathcal{I},\gamma}}\nabla_{\gamma}\log\mathbf{P}_{\mathcal{I},\gamma}(\mathcal{I})g(S,{\mathcal{I}},\text{Oracle}).
\end{aligned}
\end{equation}
Furthermore, we can take a expansion on $\mathbf{E}_{{\mathcal{I}}\sim\mathbf{P}_{\mathcal{I},\gamma}}\nabla_{\gamma}\log\mathbf{P}_{\mathcal{I},\gamma}(\mathcal{I})$ in the last line w.r.t. $\gamma=(\gamma_C,\gamma_N)$:
$$
\mathbf{E}_{{\mathcal{I}}\sim\mathbf{P}_{\mathcal{I},\gamma}}\nabla_{\gamma}\log\mathbf{P}_{\mathcal{I},\gamma}(\mathcal{I})=
\left(               
  \begin{array}{c}
    \mathbf{E}_{N\sim\mathbf{P}_{N,\gamma_N}}\mathbf{E}_{x_1,..,x_N\sim\prod_{i=1}^{N}\mathbf{P}_{C,\gamma_C}}
    \nabla_{\gamma_C}(\sum_{i=1}^{N}\log\mathbf{P}_{C,\gamma_C}(x_i))\\  
    \ \\
    \mathbf{E}_{N\sim\mathbf{P}_{N,\gamma_N}} \nabla_{\gamma_N} \mathbf{E}_{x_1,..,x_N\sim\prod_{i=1}^{N}\mathbf{P}_{C,\gamma_C}}\log\mathbf{P}_{N,\gamma_N}(N)\\ 
  \end{array}
\right)  
$$
After taking the above formula into Eq.~\ref{derive LDG}, we complete the derivation of gradients about the Data Generator.

\subsection{Computation of Log-Probability}\label{app: log prob compute}
An extra computation is needed for the log-probability in Eq.~\ref{gradient of attack1 TSP} and we do this in the following way:
Assuming it's independent between each dimension in a two-dimension coordinate, we only show the one-dimension case without loss of generality.

Formally, there are two random variables $X\sim\text{U}(0,1)$ and $Y\sim\text{N}(0,\sigma^2)$, and we are to compute the probability density function of random variable $Z=X+Y$. We get:
$$
\mathrm{P}(Z\le z) = \mathrm{P}(X+Y\le z) = \int_{0}^{1}dx\int_{-\infty}^{z-x}\frac{1}{\sqrt{2\pi}\sigma}\exp({-\frac{y^2}{2\sigma^{2}}})dy
$$
and we have:
$$
\mathrm{p}(z) =\frac{d\mathrm{P}(Z\le z)}{dz}=\int_{0}^{1}\frac{1}{\sqrt{2\pi}\sigma}\exp({-\frac{(z-x)^2}{2\sigma^{2}}})\text{d}x.
$$
All we need to do is to approximate this integration. Various methods can be used do so. In this work, we handle this simple integration by Monte Carlo sampling by sampling 10000 samples within $[0,1]$ to make a rough approximation.
After obtaining the approximated probability, we can easily get the log-probability due to the independent assumption.

\subsection{Algorithm}
We show the algorithm in Alg.~\ref{alg:PSRO for CO}.
\begin{algorithm}
  \caption{Two-Player Framework for Combinatorial Optimization}
  \label{alg:PSRO for CO}
\begin{algorithmic}
  \STATE {\bfseries Input:} Initial joint policy sets for Solver Selector and Data Generator as $\Pi$. Compute utilities $U^{\Pi}$ for joint $\pi\in\Pi$. Initialize meta-strategies $\sigma_i=\text{UNIFORM}(\Pi_i)$
  \WHILE{epoch e in $\{1,2,...\}$}
      \STATE  Construct mixing distribution $\pi_{\text{mix}}=\sum_i\sigma_{\text{DG}}^i\pi_{\text{DG}}^i$
        and train the oracle for Solver Selector $S^{'}$ with gradient in Eq.~\ref{gradient of LSS}
      \FOR{many episodes}
        \STATE Sample $S\sim\Pi_{\text{SS}}$ according to $\sigma_{\text{SS}}$
        \STATE Train Oracle $\mathbf{P}^{'}_{\mathcal{I}}=\text{br}(S)$ with gradient in Eq.~\ref{gradient of LDG}
        \ENDFOR
    \STATE Update policy set:$\Pi\leftarrow\Pi\cup\{(S^{'},\mathbf{P}^{'}_{\mathcal{I}})\}$
    \STATE Compute missing entries in $U^{\Pi}$ from $\Pi$ and  the meta-strategy $\sigma$ from $U^{\Pi}$
    \ENDWHILE
    \STATE Output meta-strategy $\sigma_{\text{SS}}$ and policy set $\Pi_{\text{SS}}$ to obtain mixing model by Eq.~\ref{Q-mixing} or Eq.~\ref{PG-mixing}.
\end{algorithmic}
\end{algorithm}

\subsection{Detailed Experimental Settings}\label{app:exp settings}
\textbf{Hyperparameters.} We don't propose any specific RL Solver in this work since our method is a unified framework to suit any previous models. So in this paper, we use \textbf{LIH} as our base model. All settings about the RL Solver is same as the original paper. 

For the settings of Data Generator, we initialize the $\gamma_N$ randomly and use a simple three-layer neural networks to represent the attack generator $f_{\gamma_C}$ in Eq.~\ref{attack sigma}. We also use a Sigmoid function as the last layer to scale the variation within $[0,1]$ and an additional scalar $\lambda\in[0,1]$ to make a further limit within $[0,\lambda]$.  Here we set $\lambda=\frac{1}{3}$ because of the '68-95-99.7 rule' which is a famous principle in statistics. It not only guarantees each point within [0,1] can reach any other point after adding a gaussian perturbation, but also makes few changes to the structure of original Instances after normalization in Eq.~\ref{normalization}.

All parameters in our framework except for those in the RL Solver are updated by Adam~\citep{kingma2014adam} optimizer with specific learning rate settings and the overall configuration of this neural networks is shown in Table~\ref{attack config}.
\begin{table}[H]
\vspace{-10mm}
\caption{Configuration of Attack Neural Networks}
\label{attack config}
\begin{center}
\begin{tabular}{ll}
\multicolumn{1}{c}{\bf Module}  &\multicolumn{1}{c}{\bf DESCRIPTION}
\\ \hline \\
First Layer        &  dim=2 with ReLU activation \\
Second Layer        &  dim=128 with ReLU activation \\
Output Layer         &dim=2 with Sigmoid activation \\
Optimizer             &Adam with lr=0.05, lr\_decay=0.95\\
$l_2$ norm        & weight\_decay=0.01\\
Epochs            & 40 for TSP20 TSP50 and TSP100
\end{tabular}
\end{center}
\vspace{-10mm}
\end{table}

During the training at each PSRO loop, we choose the attack generator where the current Solver Selector performs worst for the next PSRO loop. Similarly, we choose the Solver with best performance on the mixing distribution constructed by the current Data Generator. Specifically, we generate a validation set by sampling 1000 Instances from the distribution constructed by the Data Generator's policy set and its meta-strategy. We then test each epoch's model on this dataset and select the best one as the model to be trained in the next PSRO loop.

\textbf{Fine-tune version.} We also provide a fine-tune version under our training framework. Specifically, we pick the model as a warm start which has pretrained on the uniform distribution and continue to train them under the framework of PSRO, which can be seen as the fine-tune process to overcome the weakness of the current model. We call this version of model \textbf{LIH(FT)} in the following. Respectively, we denote the version of model that trains from scratch as \textbf{LIH(FS)}. For practical use, we often use the fine-tune model rather than the model train from scratch because of the limit of time and computational resource, and in this work, we test the fine-tune model trained on TSP100 to solve the Instances from TSPLIB~\citep{reinelt1991tsplib}).

\textbf{Setup in meta-level.} Under the framework of PSRO, we train oracles for Solver Selector and Data Generator at each PSRO loop. During training \textbf{LIH(FS)}, we set the same training epochs for RL Solver: 40 epochs (per PSRO loop) for TSP20, TSP50 and TSP100 and we train 7 PSRO loop in each case.. When we train the fine-tune version, \textbf{LIH(FT)}, we use the model in the last epoch of training period in original paper as our pretrained model (for~\citep{wu2021learning}, we use the model trained after 200 epochs.) Then we train 10 epochs for TSP100 in each PSRO loop and train 8 PSRO loop. During the training, the Solver inherit the parameters of last PSRO loop and continue to train in the new PSRO loop. Noticing that we obtain a population of Solvers by 280 epochs of training, we train 280 epochs for \textbf{LIH} rather than 200 epochs in its original paper to guarantee the fair comparison.

\textbf{Mixing-model.} After getting a population of Solvers, we use the mixing policy obtained by Eq.~\ref{Q-mixing} or \ref{PG-mixing} to combine these Solver. Considering we need to get each Solver's policy during each decision step, we need to execute forward propagation for each Solver so the running time will grow linearly if there are no implementation-level tricks. As a consequence, we only use the Solvers whose probabilities are accumulated no less than 0.99 because of Solver Selector's sparse meta-strategy.

\subsection{Meta Strategy in Different PSRO Loops}\label{app:ms}
\begin{figure}[htbp!]
\vspace{-10mm}
\centering
\subfigure[Meta-strategy of the population of Solvers trained on TSP20.]{
\label{fig:ms20}
\includegraphics[scale=.2]{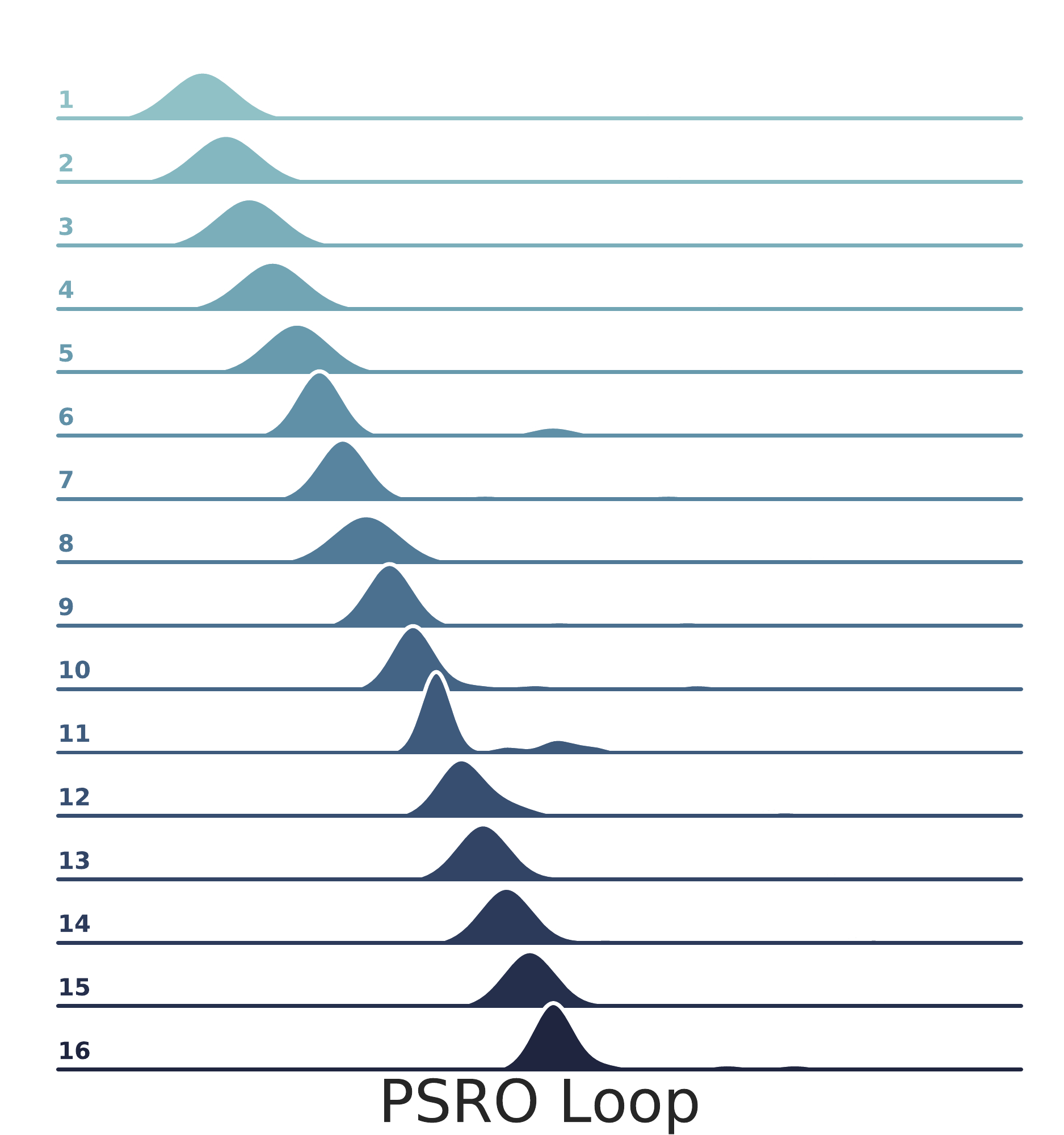}}
\hspace{.111in}
\subfigure[Meta-strategy of the population of Solvers trained on TSP50.]{
\label{fig:ms50}
\includegraphics[scale=.2]{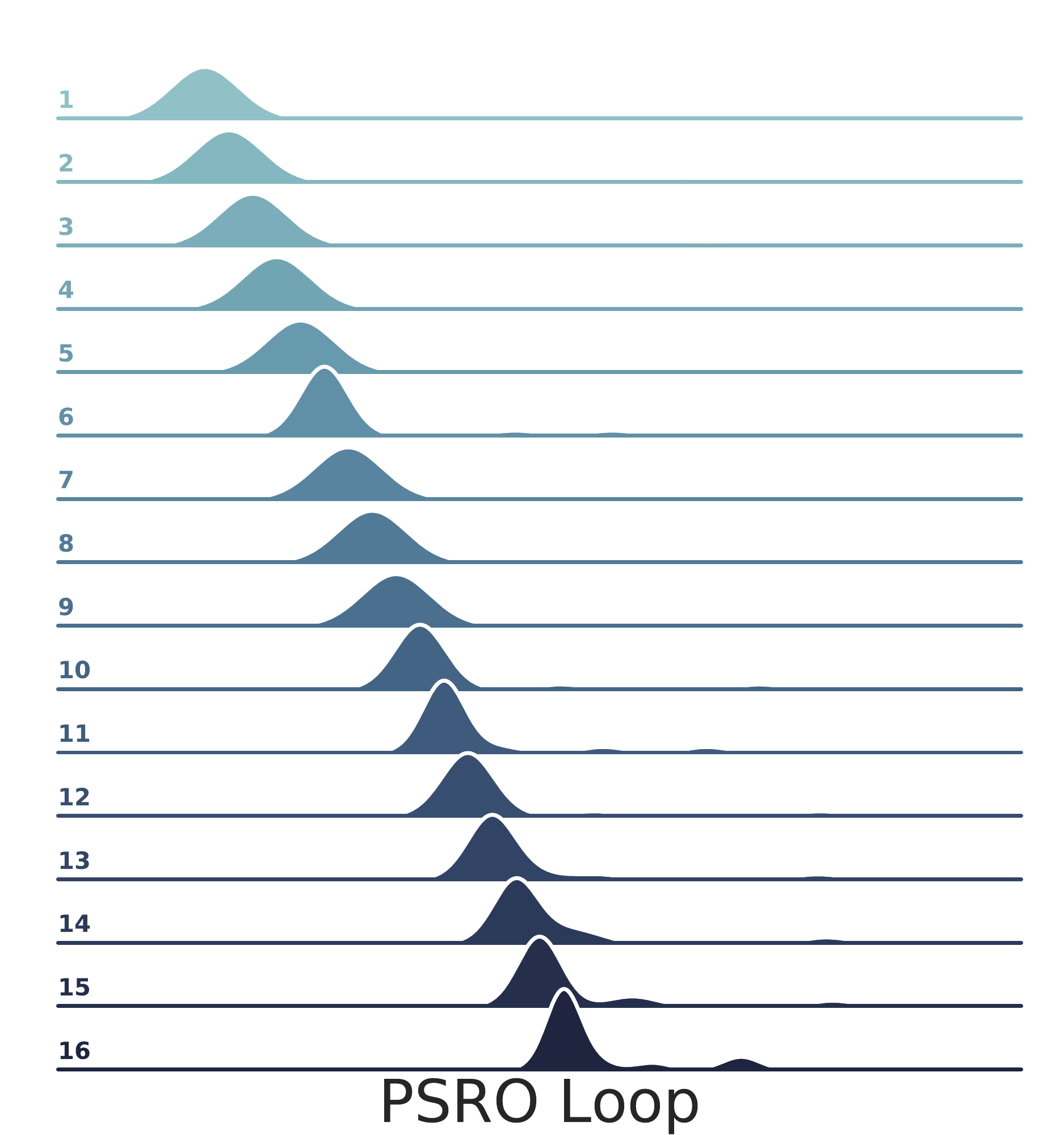}}
\hspace{.111in}
\subfigure[Meta-strategy of the population of Solvers trained on TSP100.]{
\label{fig:ms100}
\includegraphics[scale=.2]{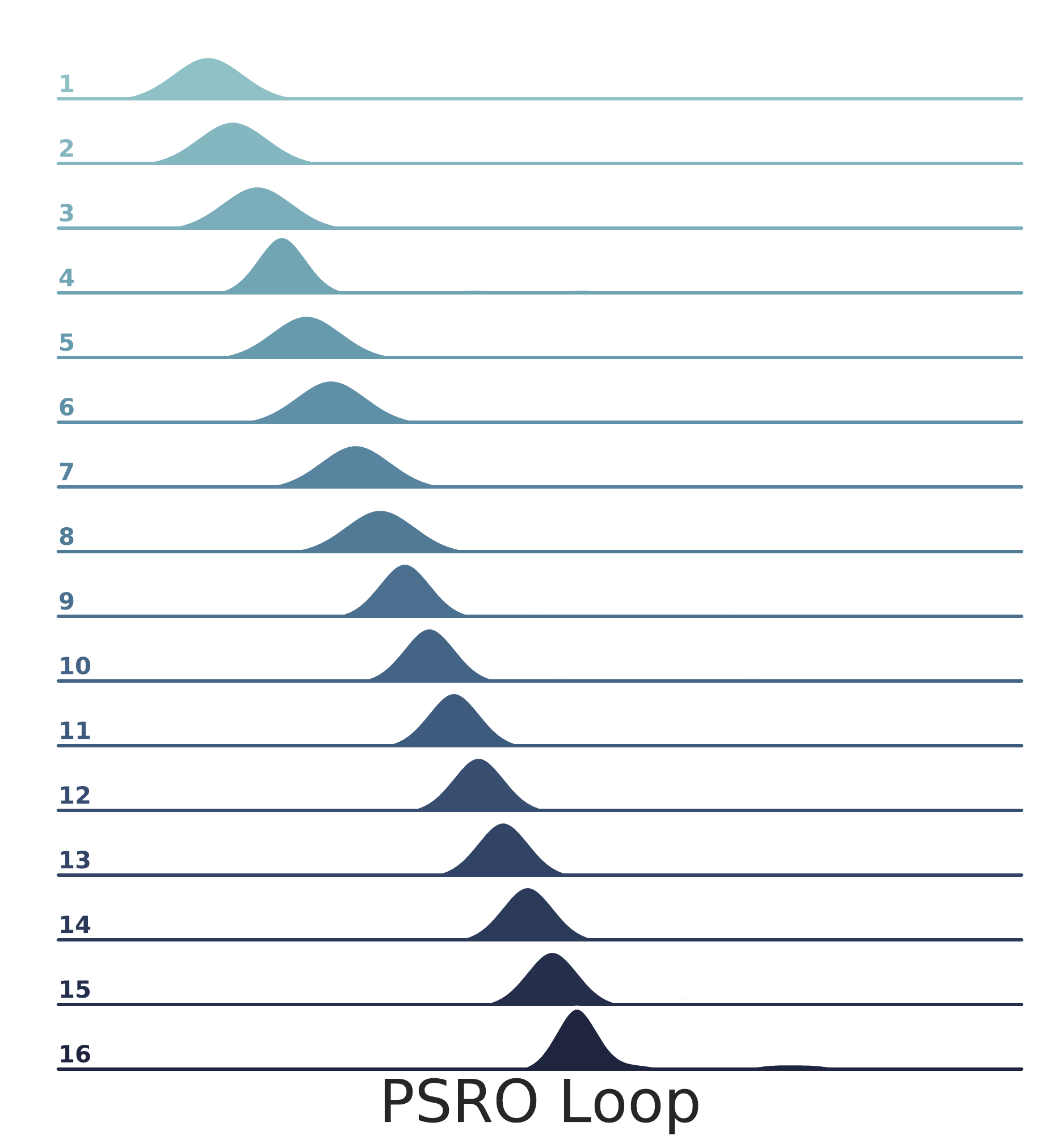}}
\vspace{-3mm}
\caption{Meta strategy of the model population.}
\label{fig:ms} 
\end{figure}

We visualize the meta strategy in every PSRO loops in Fig.~\ref{fig:ms}. Results show that at each loop, there exists the strongest Solver with a dominate meta strategy probability, leading to a quite sparse meta strategy distribution.

\subsection{Weakness of Solvers}\label{app:attack effects}
Under our framework, it's interesting to find some distributions where the Solvers (or methods) can perform poorly,  revealing the weakness of the solver. It can also provide a rough judgement on the stability of a method. We are amazed to find that only using simple multi-layer neural networks, the same as that during training oracles for Data Generator, the methods show diverse performance, as shows in Appendix.~\ref{fig:attack_LIH} and \ref{fig:attack_AM}. Therefore, it's reasonable to take this criterion into consideration when comparing different methods. However, there are few researches about the exploration about the weakness but we think it's quite important especially in realistic applications.

We demonstrate performance can be influenced a lot even by adding small gaussian perturbations in Fig.~\ref{fig:attack_LIH} and \ref{fig:attack_AM}. We use the model trained in corresponding paper: training 200 epochs for LIH~\citep{wu2021learning} and 100 epochs for AM~\citep{DBLP:conf/iclr/KoolHW19} on TSP20, TSP50 and TSP100. Results show that our attack generator can learn a distribution where the well-trained model performs bad, which motivates us to employ such method to train oracles under the framework of PSRO. 

\begin{figure}[htbp!]
\centering
\includegraphics[scale=.5]{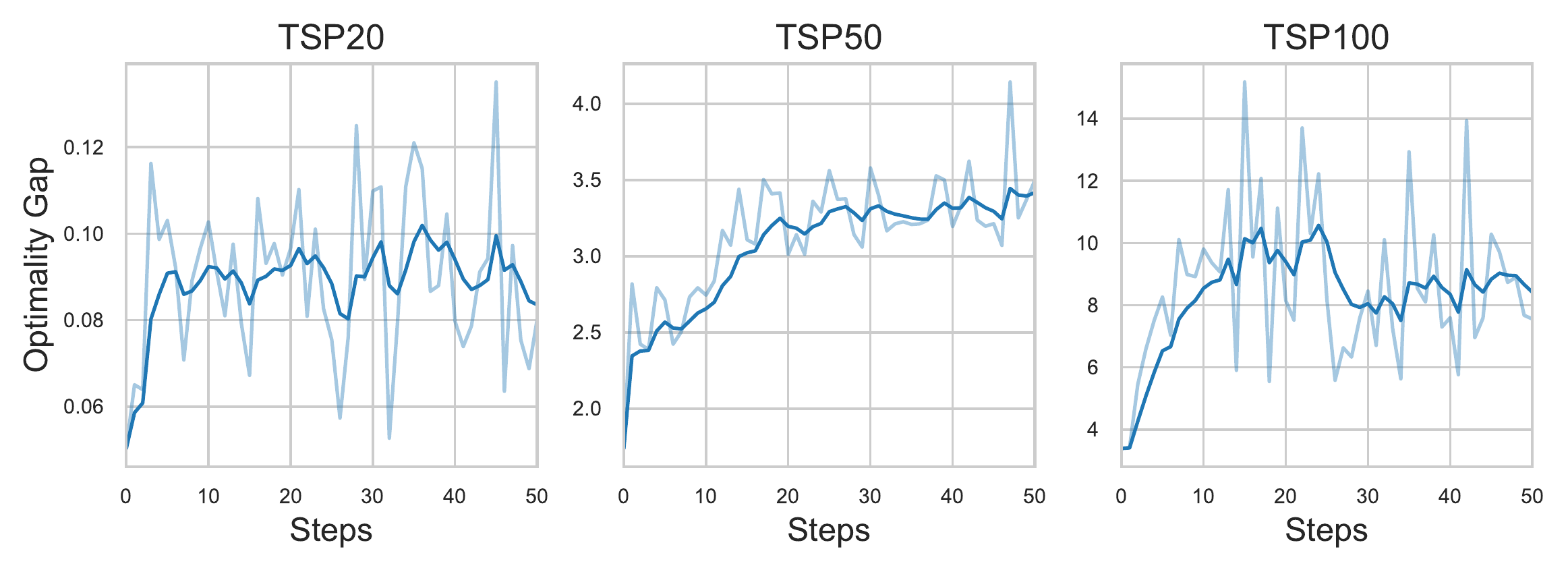}
\caption{Training figure of attack generator for LIH~\citep{wu2021learning}.}
\label{fig:attack_LIH} 
\end{figure}

\begin{figure}[htbp!]
\centering
\includegraphics[scale=.5]{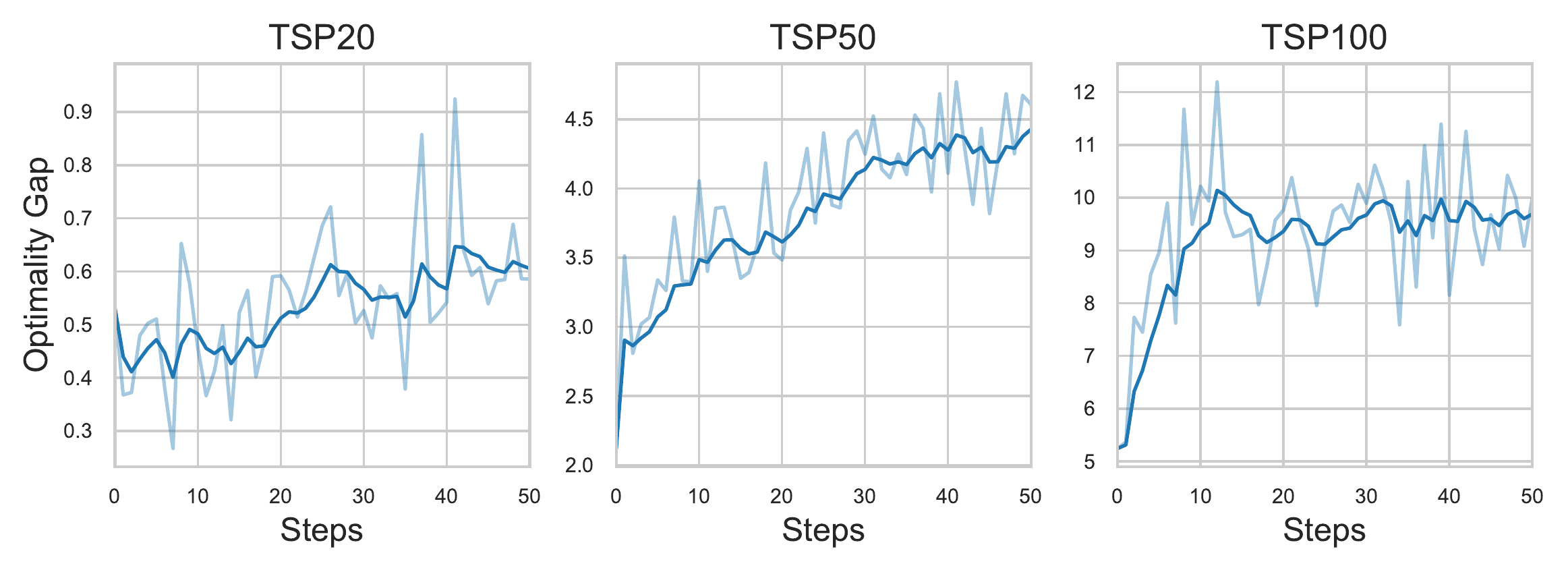}
\caption{Training figure of attack generator for AM~\citep{DBLP:conf/iclr/KoolHW19}}
\label{fig:attack_AM} 
\end{figure}

\subsection{Demonstration of Attack Distribution}
\begin{figure}[htbp!]
\centering
\subfigure[Attack distribution generated by PSRO trained on TSP20.]{
\label{fig:attack dist_20}
\includegraphics[scale=.2]{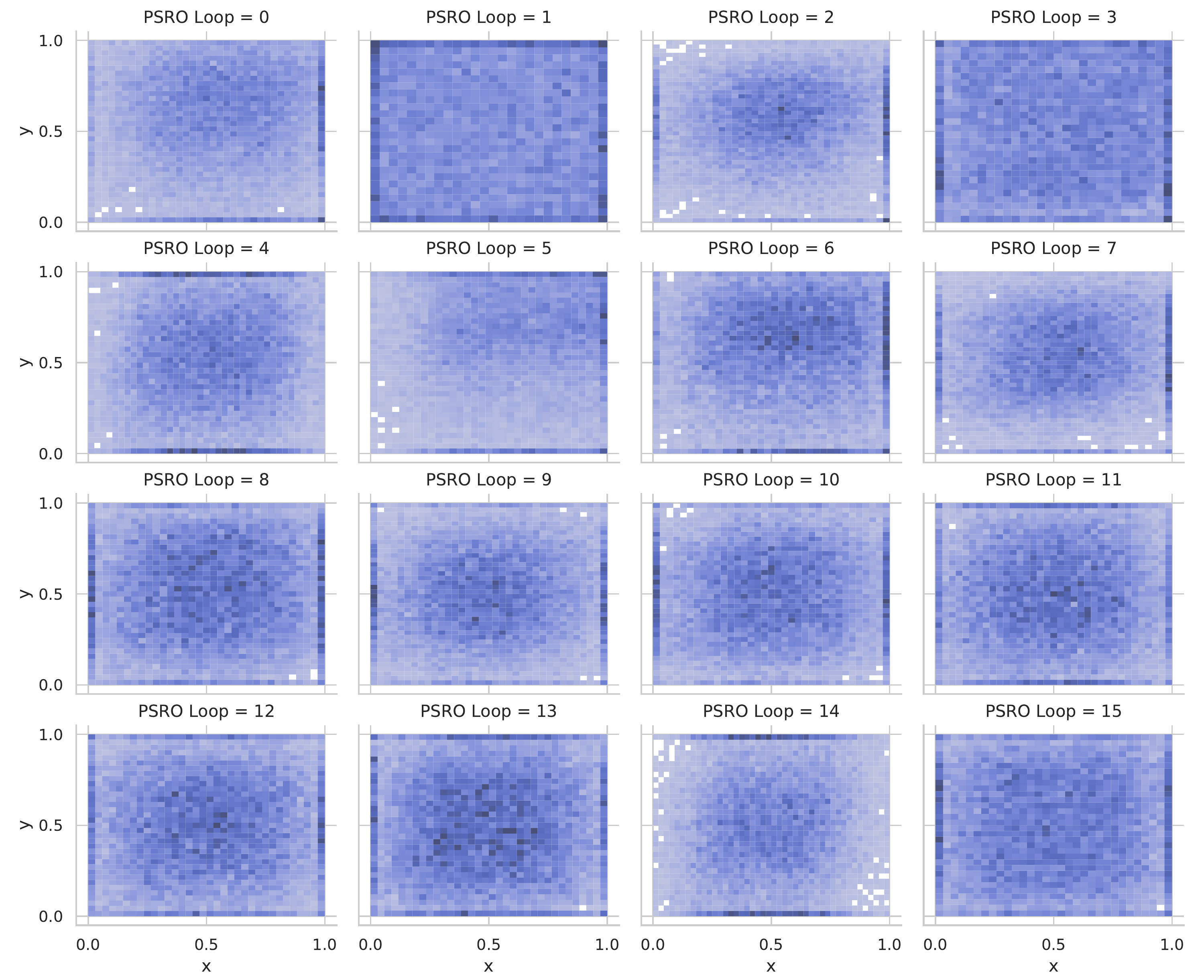}}
\hspace{2mm}
\subfigure[Kernel density estimations of attack distribution generated by PSRO trained on TSP20.]{
\label{fig:attack dist_kde_20}
\includegraphics[scale=.2]{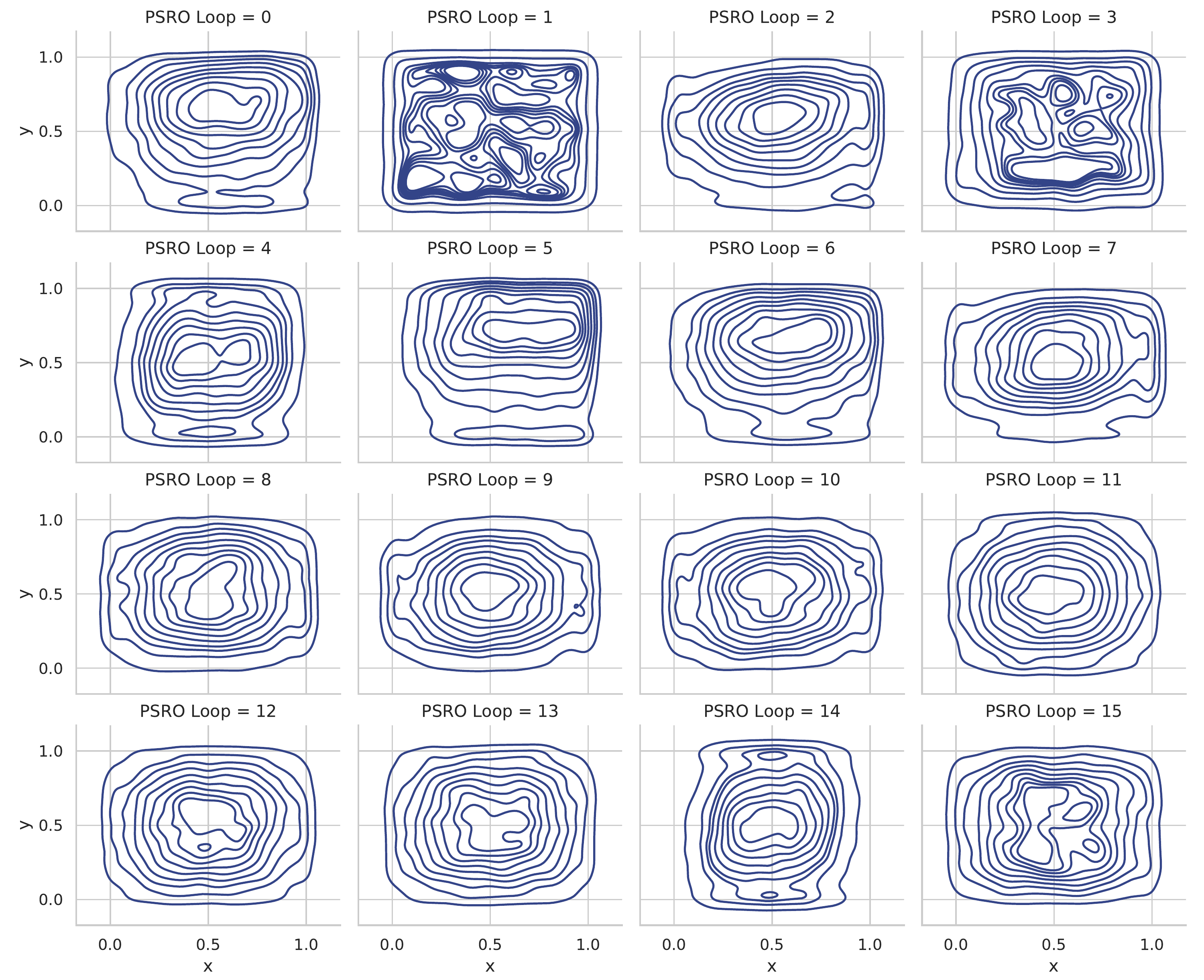}}
\vspace{2mm}
\subfigure[Attack distribution generated by PSRO trained on TSP50.]{
\label{fig:attack dist_50}
\includegraphics[scale=.2]{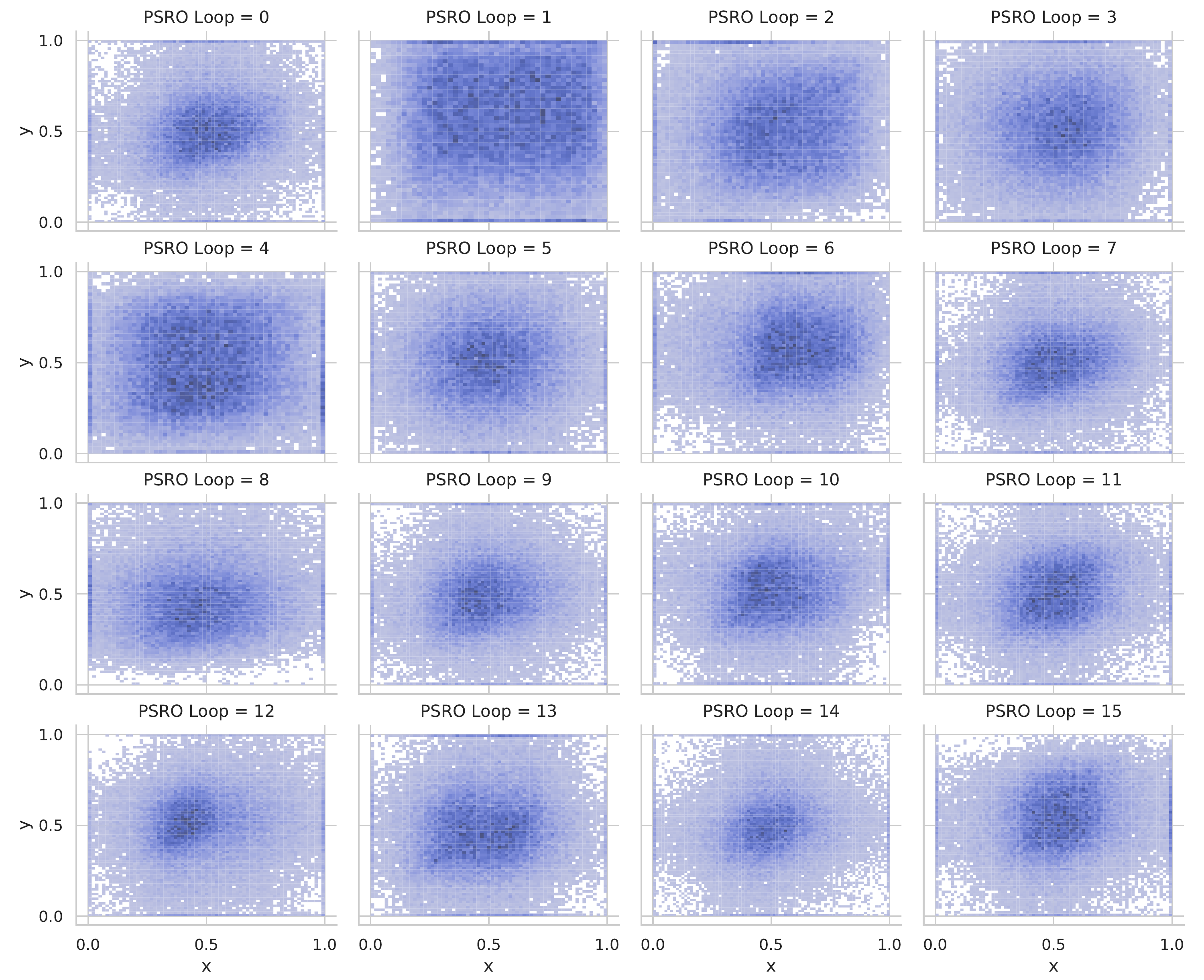}}
\hspace{2mm}
\subfigure[Kernel density estimations of attack distribution generated by PSRO trained on TSP50.]{
\label{fig:attack dist_kde_50}
\includegraphics[scale=.2]{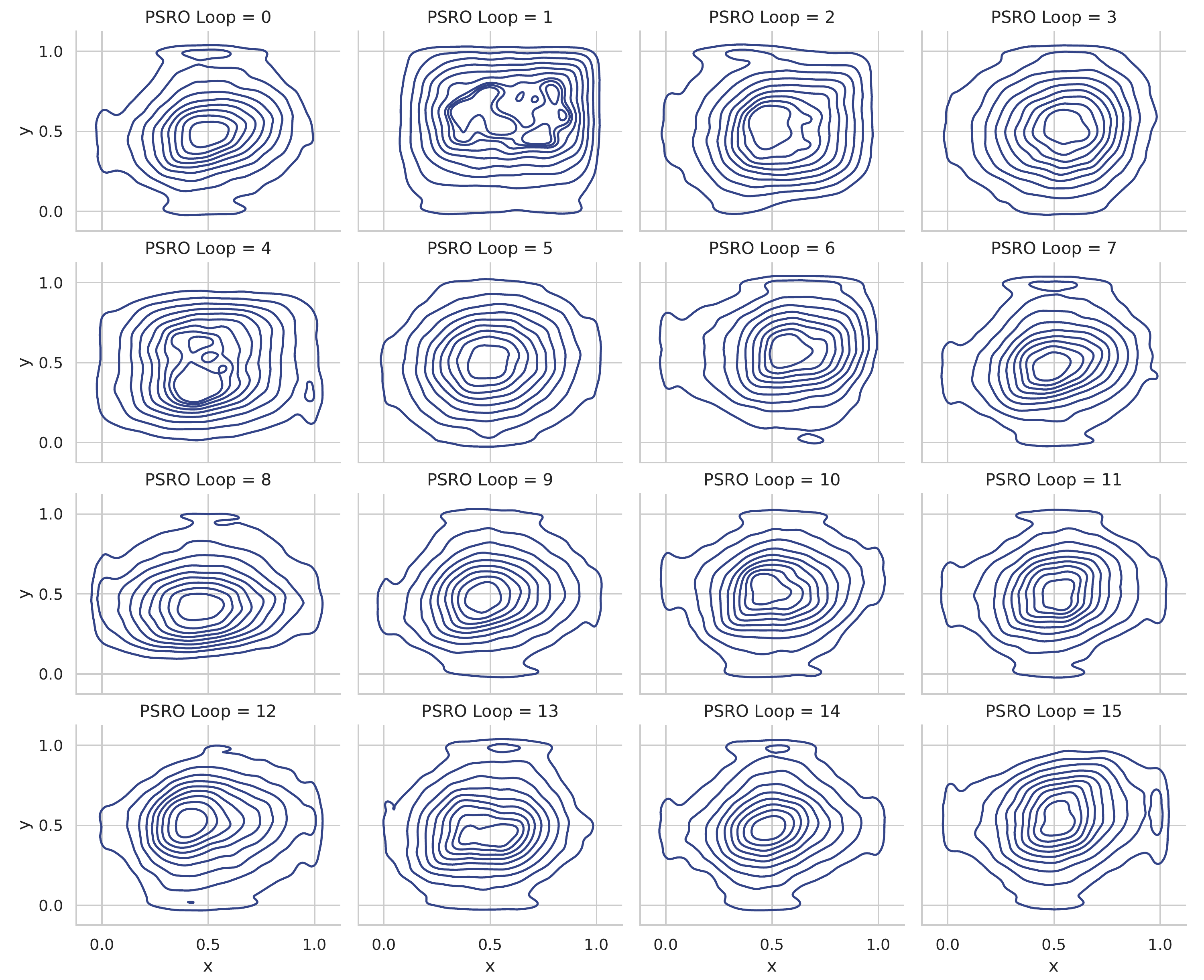}}
\vspace{2mm}
\subfigure[Attack distribution generated by PSRO trained on TSP100.]{
\label{fig:attack dist_100}
\includegraphics[scale=.2]{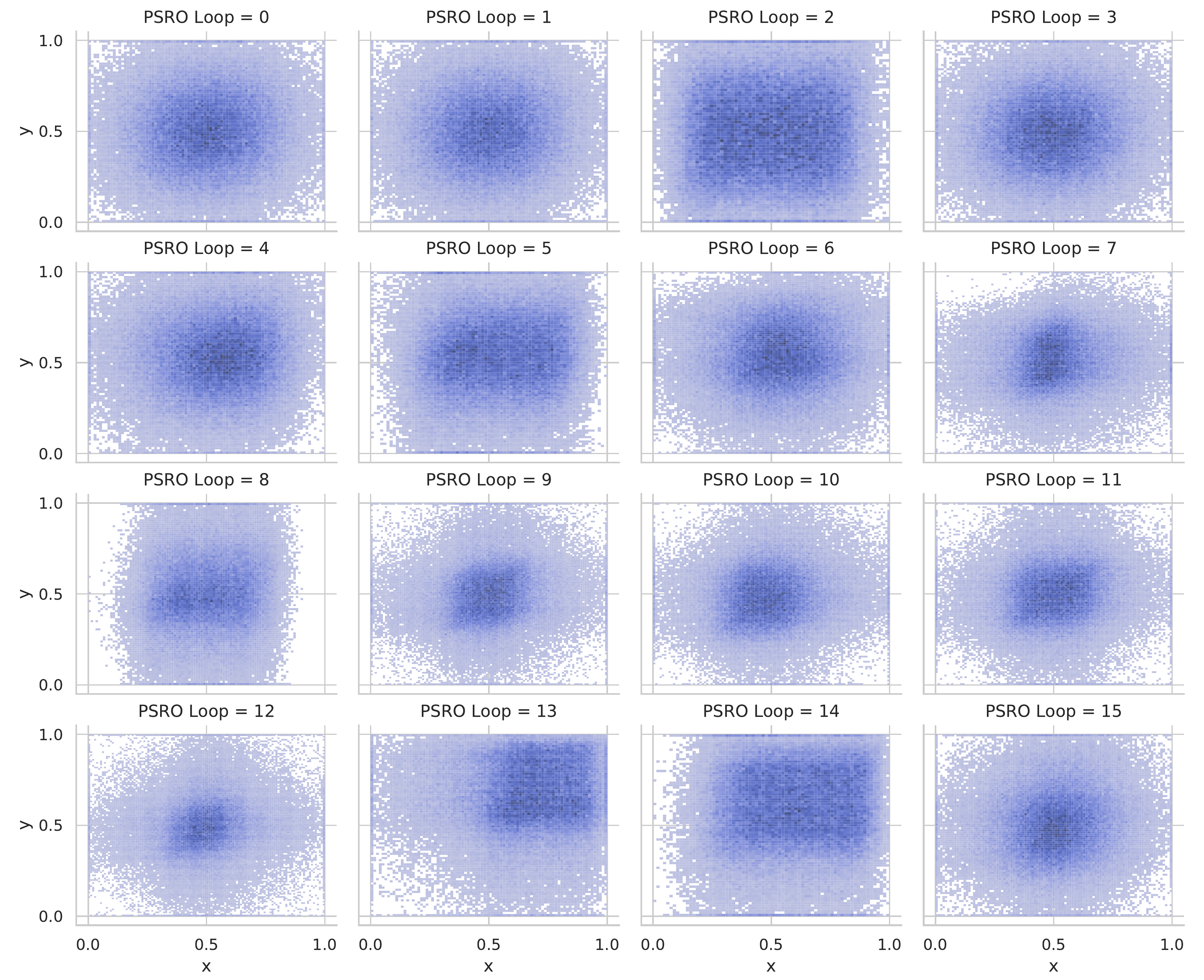}}
\hspace{2mm}
\subfigure[Kernel density estimations of attack distribution generated by PSRO trained on TSP100.]{
\label{fig:attack dist_kde_100}
\includegraphics[scale=.2]{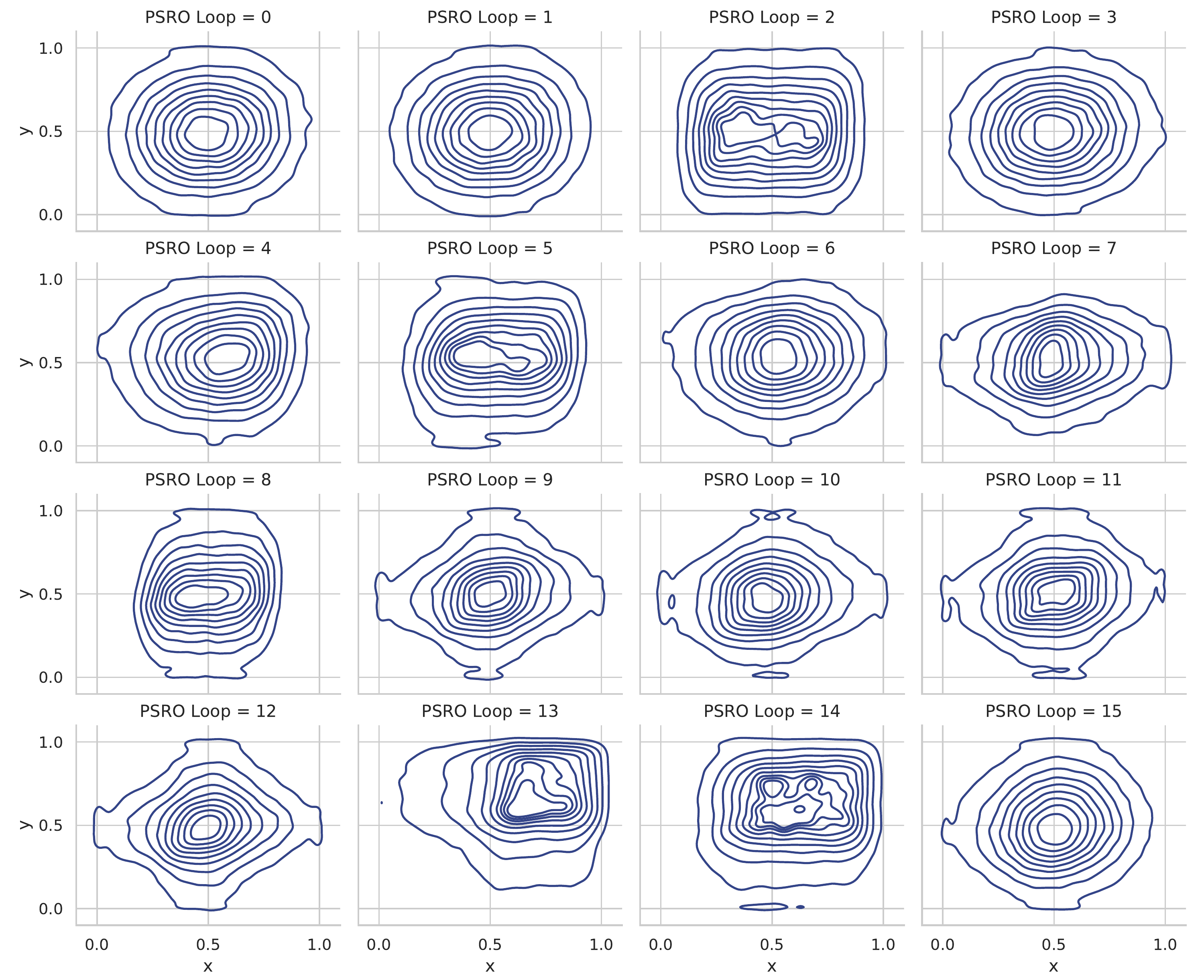}}
\caption{Attack distribution generated by PSRO}
\label{fig:attack dist}
\end{figure}

We visualize the attack distribution obtained by each PSRO loop in Fig.~\ref{fig:attack dist}. Specifically, Fig.~\ref{fig:attack dist_20},~\ref{fig:attack dist_50} and~\ref{fig:attack dist_100} are points which comprises 1000 Instances. Fig.~\ref{fig:attack dist_kde_20},~\ref{fig:attack dist_kde_50} and~\ref{fig:attack dist_kde_100} are corresponding kernel density estimations.

\end{document}

%% file: math_commands.tex
\usepackage{amsmath,amsfonts,bm}

\def\eqref#1{equation~\ref{#1}}

\def\1{\bm{1}}

\DeclareMathAlphabet{\mathsfit}{\encodingdefault}{\sfdefault}{m}{sl}
\SetMathAlphabet{\mathsfit}{bold}{\encodingdefault}{\sfdefault}{bx}{n}

